\documentclass[times, review, 10pt]{elsarticle}

\usepackage{hyperref}

\usepackage{acro}
\usepackage{amsmath,amssymb,mathrsfs}
\usepackage{booktabs}
\usepackage{tikz}
\usepackage{multirow}
\usetikzlibrary{positioning}
\usetikzlibrary{calc} 
\usepackage[final]{microtype}
\usepackage{subcaption}
\usepackage{pdflscape}
\usepackage{enumitem}
\usepackage{rotating}
\usepackage{pdflscape}

\DeclareAcronym{op}{
  short = OP ,
  long  = Osteoporosis ,
  sort  = OP ,
}
\DeclareAcronym{oa}{
  short = OA ,
  long  = Osteoarthritis ,
  sort  = OA ,
}
\DeclareAcronym{msk}{
  short = MSK ,
  long  = Musculoskeletal ,
  sort  = MSK ,
}
\DeclareAcronym{cnn}{
  short = CNN ,
  long  = Convolutional Neural Network ,
  sort  = CNN ,
}
\DeclareAcronym{ao}{
  short = AO ,
  long  = Adaptive Optics ,
  sort  = AO ,
}
\DeclareAcronym{dnn}{
  short = DNN ,
  long  = Deep Neural Network ,
  sort  = DNN ,
}
\DeclareAcronym{psf}{
  short = PSF ,
  long  = Point Spread Function ,
  sort  = PSF ,
}
\DeclareAcronym{mse}{
  short = MSE ,
  long  = Mean Squared Error ,
  sort  = MSE ,
}
\DeclareAcronym{rmse}{
  short = RMSE ,
  long  = Root Mean Squared Error ,
  sort  = RMSE ,
}
\DeclareAcronym{ddpm}{
  short = DDPM ,
  long  = Diffusion-Based Denoising Probabilistic Model ,
  sort  = DDPM ,
}
\DeclareAcronym{gan}{
  short = GAN ,
  long  = Generative Adversarial Network ,
  sort  = GAN ,
}

\DeclareAcronym{mlp}{
  short = MLP ,
  long  = Multi-Layer Perceptron ,
  sort  = MLP ,
}
\DeclareAcronym{psnr}{
  short = PSNR ,
  long  = Peak Signal to Noise Ratio ,
  sort  = PSNR ,
}
\DeclareAcronym{ssim}{
  short = SSIM ,
  long  = Structural Similarity Index ,
  sort  = SSIM ,
}

\DeclareAcronym{gnn}{
  short = GNN ,
  long  = Graph Neural Network ,
  sort  = GNN ,
}
\DeclareAcronym{3d}{
  short = 3D ,
  long  = 3-dimensional ,
  sort  = 3D ,
}
\DeclareAcronym{zrnet}{
  short = ZRNet ,
  long  = Zernike coefficient prediction and optical image Restoration Network ,
  sort  = ZRNet ,
}

\DeclareAcronym{faa}{
  short = FAA ,
  long  = Frequency-aware Alignment ,
  sort  = FAA ,
}

\DeclareAcronym{rmswfe}{
  short = RMS\textsubscript{WFE} ,
  long  = Root Mean Square Wavefront Error,
  sort  = RMS\textsubscript{WFE} ,
}
\DeclareAcronym{sota}{
  short = SOTA ,
  long  = state-of-the-art,
  sort  = SOTA ,
}
\DeclareAcronym{gat}{
  short = GAT ,
  long  = Graph Attention,
  sort  = GAT ,
}
\journal{Pattern Recognition}

\begin{document}

\begin{frontmatter}

\title {Physics-Informed Graph Neural Networks for Frequency-Aware Optical Aberration Correction}  

\affiliation[1]{organization={School of Computer Science, University of Nottingham},
    city={Nottingham},
    postcode={NG8 1BB},
    state={Nottinghamshire},
    country={United Kingdom}}

\affiliation[2]{organization={Optics and Photonics Group, Department of Electrical and Electronic Engineering, University of Nottingham},
    city={Nottingham},
    postcode={NG8 1BB},
    state={Nottinghamshire},
    country={United Kingdom}}

\affiliation[3]{organization={Research Center for Humanoid Sensing, Zhejiang Laboratory},
    city={Hangzhou},
    postcode={3111100},
    state={Zhejiang},
    country={China}}


\author[1]{Yong En Kok}

\author[1]{Bowen Deng}

\author[2]{Alexander Bentley}

\author[1]{Andrew J. Parkes}

\author[2,3]{Michael G. Somekh}

\author[2]{Amanda J. Wright}

\author[1]{Michael P. Pound \corref{cor1}}
\ead{michael.pound@nottingham.ac.uk}

\cortext[cor1]{Corresponding author}

\begin{abstract}
Optical aberrations significantly degrade image quality in microscopy, particularly when imaging deeper into samples. These aberrations arise from distortions in the optical wavefront and can be mathematically represented using Zernike polynomials. Existing methods often address only mild aberrations on limited sample types and modalities, typically treating the problem as a black-box mapping without leveraging the underlying optical physics of wavefront distortions. We propose ZRNet, a physics-informed framework that jointly performs Zernike coefficient prediction and optical image Restoration. We contribute a Zernike Graph module that explicitly models physical relationships between Zernike polynomials based on their azimuthal degrees—ensuring that learned corrections align with fundamental optical principles. To further enforce physical consistency between image restoration and Zernike prediction, we introduce a Frequency-Aware Alignment (FAA) loss, which better aligns Zernike coefficient prediction and image features in the Fourier domain. Extensive experiments on CytoImageNet demonstrates that our approach achieves state-of-the-art performance in both image restoration and Zernike coefficient prediction across diverse microscopy modalities and biological samples with complex, large-amplitude aberrations. We further validate on experimental PSF data from a physical microscope and demonstrate robustness to realistic sensor noise, confirming generalisation beyond simulated conditions. Code is available at \url{https://github.com/janetkok/ZRNet}.
\end{abstract}


\begin{highlights}
    \item Physics-informed framework for joint image restoration and aberration prediction.
    \item Zernike Graph models physical relationships between Zernike modes based on azimuthal degrees.
    \item Spatial frequency-aware loss enforces consistency in the Fourier domain.
    \item Outperforms SOTA on CytoImageNet with diverse samples, large-amplitude aberrations.
    \item Validated on experimental PSF data and realistic noise conditions.
\end{highlights}


\end{frontmatter}

\section{Introduction}
\label{sec:intro}
\begin{figure}[t]
  \centering
  \setlength{\tabcolsep}{0.8pt} 
  \begin{tabular}{ccccc} 
    \multicolumn{5}{c}{\textbf{\small Aberrated images}} \\ 
    \noalign{\vspace{1pt}} 

    \includegraphics[width=0.19\linewidth]{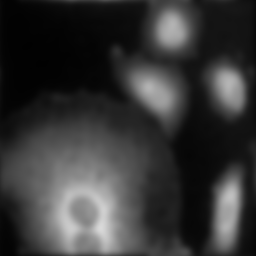} &
    \includegraphics[width=0.19\linewidth]{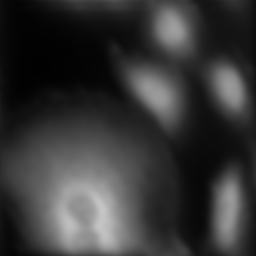} &
    \includegraphics[width=0.19\linewidth]{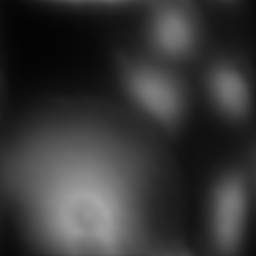} &
    \includegraphics[width=0.19\linewidth]{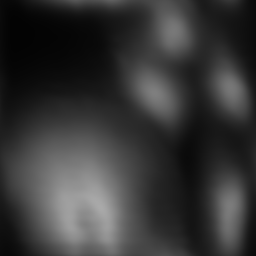} &
    \includegraphics[width=0.19\linewidth]{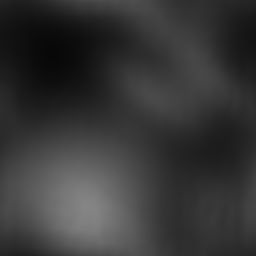} \\[-8pt] 
    
    \begin{minipage}[t]{0.19\linewidth}\centering\scriptsize{Guo et al. 2025}\end{minipage} & 
    \begin{minipage}[t]{0.19\linewidth}\centering\scriptsize{Kang et al. 2023}\end{minipage} & 
    \begin{minipage}[t]{0.19\linewidth}\centering\scriptsize{Qiao et al. 2024}\end{minipage} & 
    \begin{minipage}[t]{0.19\linewidth}\centering\scriptsize{Krishnan et al. 2020}\end{minipage} &
    \begin{minipage}[t]{0.19\linewidth}\centering\scriptsize{Ours}\end{minipage} \\[8pt] 

    \multicolumn{1}{c}{} & 
    \includegraphics[width=0.19\linewidth]{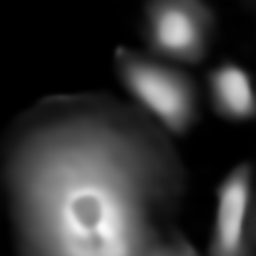} &
    \includegraphics[width=0.19\linewidth]{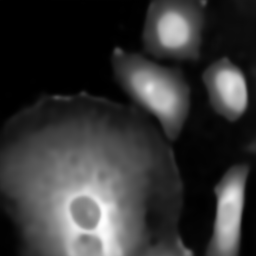} &
    \includegraphics[width=0.19\linewidth]{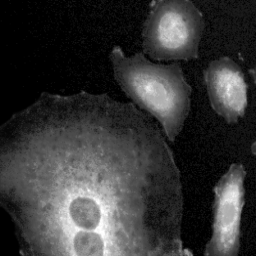} &
    \multicolumn{1}{c}{} \\[-8pt] 
    
    \multicolumn{1}{c}{} & 
    \begin{minipage}[t]{0.19\linewidth}\centering\scriptsize{Corrected (Baseline)}\end{minipage} & 
    \begin{minipage}[t]{0.19\linewidth}\centering\scriptsize{Corrected (Ours)}\end{minipage} & 
    \begin{minipage}[t]{0.19\linewidth}\centering\scriptsize{GT}\end{minipage} &
    \multicolumn{1}{c}{} \\
  \end{tabular}

  \caption{Comparison of aberration severity and correction performance. The top row shows aberrated images: those generated using implementations from previous works and an aberrated image from our approach. Our method applies approximately twice as many Zernike coefficients with larger amplitudes, resulting in a more severely aberrated image. The bottom row displays the corresponding corrected images from the baseline method and our method, respectively, along with the ground truth for reference.}
  \label{fig:aberration_considered}
\end{figure}

High-resolution optical imaging is essential across a variety of fields, particularly medical, astronomical and microscopic imaging \cite{hampson2021adaptive}. However, optical imaging quality is often degraded by system and sample-induced wavefront aberrations. System-induced aberrations arise from manufacturing or assembly imperfections, whereas sample-induced aberrations stem from variations in refractive index of the specimen \cite{schwertner2004measurement}. To improve optical imaging quality, \ac{ao} has emerged as a powerful technique to correct these aberrations, in which the aberration is first quantified before a corrective phase pattern is applied to the wavefront. Aberrations are typically represented as a weighted linear combination of Zernike polynomials, and the challenge of accurately determining which Zernikes are present when imaging deep into a sample has existed for decades. While traditional \ac{ao} approaches using a wavefront sensor \cite{platt2001history} offer high-speed aberration measurement, they typically require a known reference point in the sample, complicating preparation and limiting applicability to complex specimens where introducing exogenous references is impractical \cite{tao2011adaptive}. Sensorless \ac{ao} has emerged as a cost-effective alternative, advancing from classical iterative optimisation methods, such as stochastic optimisation \cite{zommer2006simulated}, to more recent deep learning approaches. 

Deep learning methods have shown promise in directly reconstructing aberration phases \cite{wang2021deep} and recovering the Zernike coefficients \cite{hu2023universal,rai2023deep} from input aberrated intensity images. However, these methods still require an additional step of applying corrective aberrations to the optical system or post-processing through deconvolution algorithm \cite{fish1995blind}, which can introduce artifacts due to inaccurate \ac{psf} \cite{tyson2022principles} estimation. While parallel work in direct image restoration \cite{guo2025deep} offers instant visualisation, it risks generating hallucinations in heterogeneous specimens. Thus, researchers have begun to shift toward hybrid approaches that combine deep learning with physical priors that enhance correction reliability  \cite{krishnan2020optical,kang2024coordinate,qiao2024deep}.

The challenge increases substantially when imaging deeper into samples, where aberrations become more complex and severe. Deeper tissue imaging introduces compound aberrations where different Zernike modes become strongly coupled, necessitating both a larger number of modes for correction and greater amplitude compensation \cite{schwertner2004simulation,wang2015direct}. Critically, most existing deep learning methods in this field \cite{krishnan2020optical,kang2024coordinate,qiao2024deep}  are limited to addressing low-order Zernike modes at relatively small amplitudes. While a useful proof of concept, this underestimates the size of aberrations in tissue samples \cite{booth2014adaptive}. Imaging deep remains a challenge in optical microscopy, with many instruments struggling to image beyond a few hundred micrometers deep and overcoming this problem would transform the field. In contrast, our work tackles substantially more complex cases by handling roughly double the number of Zernike modes at approximately twice the amplitude of previous approaches (see Fig. \ref{fig:aberration_considered} for an illustration of the amount of aberration applied in each work). Furthermore, most existing approaches typically focus on single imaging modalities and specific cell types, leaving their generalisability to different microscopy techniques, multiple objects in the field of view, and diverse biological samples uncertain. We instead evaluate our work on CytoImageNet dataset \cite{hua2021cytoimagenet}, a large‐scale compilation of 40 publicly available microscopy datasets spanning multiple imaging modalities and varied biological classes (e.g., nuclei, mitochondria, etc.).

In natural images, \ac{sota} architectures have excelled in tasks such as denoising \cite{zamir2022restormer}, deblurring \cite{zamir2021multi,kupyn2019deblurgan,xia2023diffir}, and super-resolution \cite{lu2023virtual}. These methods have proven effective, but typically tackle simpler, more uniform distortions (e.g., motion or defocus blur) and seldom account for the complex interactions of optical distortions across the image field.

In this work, we propose a physics-informed framework that jointly solves the problem of Zernike coefficient prediction and optical image restoration. We integrate a novel Zernike \ac{gnn} that explicitly models the physical relationship between groups of Zernike coefficients, enabling the model to learn an appropriate correction. By jointly optimising the image restoration network and Zernike graph module for accurate Zernike prediction, we impose physics-based constraints that guide the network to couple denoising with aberration correction, leveraging the underlying physics that governs how different Zernike modes interact and influence image quality.

We further introduce the \ac{faa} loss, which operates in the Fourier domain where optical aberrations distinctly manifest as structured modifications to the spatial frequency content. The loss enforces consistency between image restoration and Zernike prediction through three complementary terms, ensuring that the predicted coefficients correspond to actual restorative transformations rather than dataset-specific artifacts. Our approach bridges the gap between image restoration and aberration prediction while promoting more reliable and generalisable solutions.

We demonstrate \ac{sota} performance on a diverse range of microscopy images against both leading image restoration models, and leading aberration correction models. Our primary contributions are:

\begin{itemize} 
    \item We propose \ac{zrnet}, a physics-informed framework that unifies image restoration and aberration estimation. We introduce a novel Zernike Graph Neural Network that explicitly encodes the non-linear coupling between Zernike modes based on their azimuthal degrees, ensuring the learned latent representations adhere to physical optical principles.
    \item We formulate a \ac{faa} loss function to enforce consistency between the restoration and prediction branches. By operating in the Fourier domain, this mechanism penalises phase and amplitude distortions, effectively bridging the domain gap between pixel-wise restoration and Zernike coefficient regression.
    \item We demonstrate that our joint optimisation strategy significantly improves generalisation across varied biological samples and microscopy modalities. Extensive evaluations on CytoImageNet show that \ac{zrnet} achieves \ac{sota} performance, specifically in correcting high-order, large-amplitude aberrations where previous methods typically fail.
\end{itemize}

\section{Related Work}
\label{sec:relatedwork}
\subsection{Image Restoration Networks.} 
Image restoration aims to recover high-quality images from degraded inputs. Deep learning approaches have revolutionised this field with advances in deblurring \cite{zamir2021multi,kupyn2019deblurgan}, denoising \cite{zamir2022restormer}, and super-resolution \cite{behjati2023single} tasks, enabling reliable recovery of image features across diverse degradation conditions. For a comprehensive survey of deep learning-based image restoration methods, we refer readers to \cite{su2022survey}.
Modern image restoration approaches predominantly employ encoder-decoder architectures \cite{zamir2022restormer,xia2023diffir}, which learn compressed representations while preserving fine detail through skip connections \cite{li2020underwater}. To further enhance restoration quality, various architectural innovations have been proposed, including channel attention \cite{yang2021image}, kernel-based attention \cite{zhang2023kbnet}, and multi-stage refinement \cite{zamir2021multi}.

More recently, generative networks such as \ac{gan}s \cite{kupyn2019deblurgan} and diffusion models \cite{xia2023diffir} have been developed to more effectively handle complex degradations. These frameworks learn the underlying image distribution to enable higher quality restorations through either adversarial training or iterative denoising respectively. Despite producing visually appealing results, \ac{gan}s are prone to mode collapse and training instability, while diffusion models face challenges with computational overhead and error accumulation during iterative sampling steps \cite{liu2024residual}. Transformer-based architectures have also been applied to good effect \cite{zamir2022restormer}, leveraging their ability to model long-range dependencies and global context through self-attention. 

While these architectural innovations effectively handle common image degradations such as defocus blur, motion blur, and Gaussian noise — these distortions tend to be more uniform and predictable. In contrast, optical aberrations pose unique challenges through intricate, depth-dependent, and spatially varying wavefront distortions that interact in complex ways across the image field. Robust aberration correction demands simultaneous understanding of wavefront behaviour and advanced restoration techniques, motivating the integration of sophisticated deep learning techniques with physics-informed frameworks for aberration correction.
\begin{landscape}
\begin{figure*}
\centering

   \includegraphics[width=\linewidth]{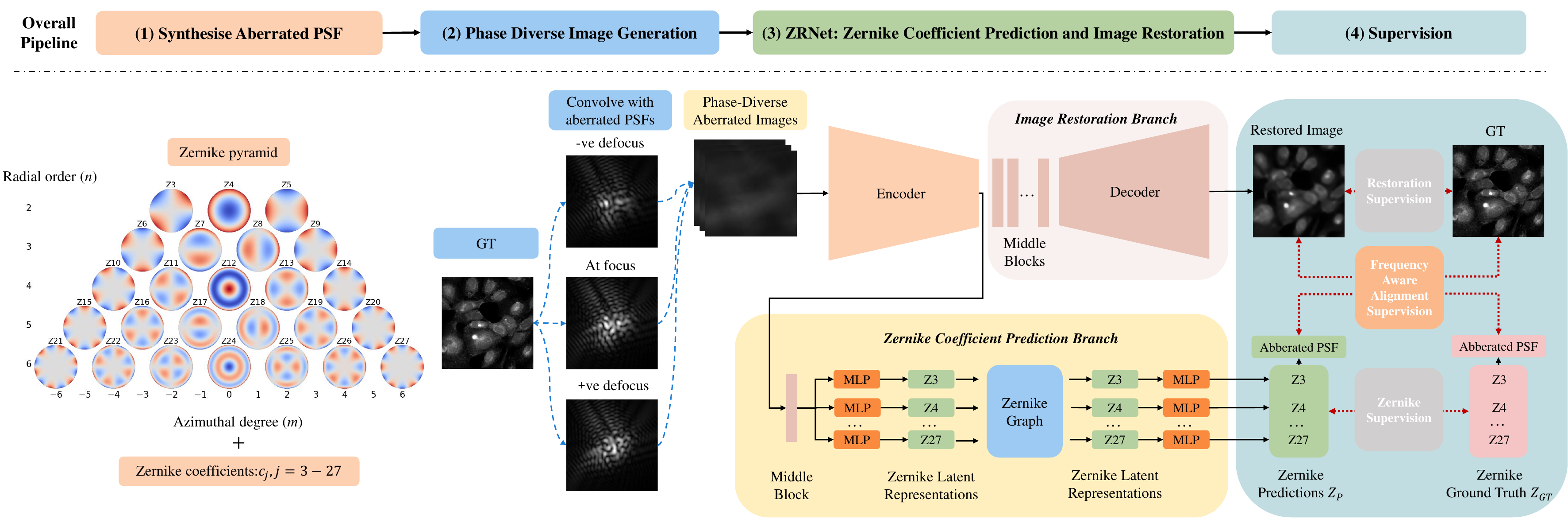}
   
   \caption{Overall pipeline: (1) synthesis of aberrated \ac{psf}s using Zernike polynomials, (2) generation of phase-diverse images by convolving ground truth samples with aberrated \ac{psf}s across multiple focal planes, (3) ZRNet that simultaneously restores images and predicts Zernike coefficients in a single forward pass, (4) with supervision across three domains: image space, frequency space, and Zernike coefficients.}
   \label{fig:overall_pipeline}
\end{figure*}
\end{landscape}
\subsection{Direct Optical Aberration Correction using Deep Learning.} 
Research on direct optical aberration correction using deep learning is less established. While approaches such as Guo et al. \cite{guo2025deep} demonstrate the effectiveness of solely using deep learning architectures for direct mapping between degraded and high-quality optical images, their method showed negligible improvement beyond 4th order Zernikes and risked hallucinations with heterogeneous specimens. This has prompted researchers to shift toward hybrid approaches that combine deep learning with physical priors to enhance correction reliability through consistency loss mechanisms. For instance, Krishnan et al. \cite{krishnan2020optical} implemented a dual-network architecture that simultaneously optimises image restoration and Zernike coefficient estimation, but struggled with higher-order aberrations and showed no clear benefits from joint training and the consistency loss. Similar consistency mechanisms have been adopted by Kang et al. \cite{kang2024coordinate}, who developed a self-supervised algorithm using coordinate-based neural representations to jointly estimate Zernike coefficients and recover 3-dimensional structure in widefield microscopy images, but this work only addressed 1-3 Zernike coefficients. Qiao et al. \cite{qiao2024deep} proposed a two-stage architecture where a dual-branch network, consisting of spatial and Fourier domain feature extraction paths, first estimates optical aberrations, with this aberration estimation then serving as input to guide a separate aberration-aware super-resolution network for image restoration. The authors observed that incorporating higher-order aberrations significantly degraded performance, even at the aberration estimation stage, which could potentially misdirect the subsequent image restoration process.

While progress has been made, image restoration of large amplitude aberration, common in tissue samples at increasing depth, has yet to be solved. Our work addresses this challenge, with higher-order aberrations (up to 6th order, 25 Zernike coefficients) at larger amplitudes. Moreover, we uniquely incorporate explicit modelling of the physical relationships between aberration modes, providing valuable constraints for more accurate estimation and correction. This broader scope and deeper physical integration represents a significant departure from previous approaches, tackling more challenging optical aberration scenarios.

\section{Aberration Dataset}
\label{sec:aberrationdataset}
\subsection{Aberrated Point Spread Function (PSF).} 
Optical aberrations in an imaging system, such as a microscope, can be modeled using its point spread function (\ac{psf}). The \ac{psf} describes how the imaging system responds to a single point source of light and represents the system’s spatial resolution. In an ideal system, the \ac{psf} is a diffraction-limited single point of finite size and shape, while aberrations in real systems further distort it \cite{Goodman2005Fourier}, degrading image quality. The \ac{psf} is calculated by taking the Fourier transform of the complex pupil function, which defines how light is transmitted through an optical system:
\begin{equation}
PSF(x,y) = \left|\mathcal{F}\{A(x,y) \exp\left(i\phi(x,y)\right)\}\right|^2
  \label{eq:psf}
\end{equation}
where, $A(x,y)$ is a circular aperture mask (1 inside, 0 outside), $\phi$ is the wavefront phase and $\mathcal{F}$ is the Fourier transform. 

To model the wavefront phase $\phi$, we use Zernike polynomials, which provide a complete mathematical basis for describing wavefront aberrations. These polynomials can be defined as either odd or even functions:
\begin{equation}
Z^m_n(\rho,\phi)=R^m_n(\rho)cos(m\phi)
\label{eq:zernike_odd}
\end{equation}
and odd polynomials given by:
\begin{equation}
Z^{-m}_n(\rho,\phi)=R^m_n(\rho)sin(m\phi)
\label{eq:zernike_even}
\end{equation}
where $R^m_n$ is a radial polynomial dependent on m, n and the radial distance $\rho$ and $\phi$ is the azimuthal angle. Zernike polynomials are particularly useful because they are orthogonal over a unit circle \cite{noll1976zernike} and can be grouped by radial order ($n$), azimuthal degree ($m$), or aberration type to collectively model different types of optical distortions (see Zernike pyramid in Fig. \ref{fig:overall_pipeline}).

In this work, the wavefront phase $\phi$ is defined through the linear combination of weighted Zernike polynomials 3-27 (ANSI indices), $Z_j$ in Eqn. \ref{eq:phi}, excluding $Z_0$ (piston), $Z_1$ (tip), and $Z_2$ (tilt) modes as they can be corrected using standard centroiding or registration methods.
\begin{equation}
\phi(x,y) = \sum_{j=3}^{27} c_j Z_j(x,y)
\label{eq:phi}
\end{equation}

where $c_j$ are coefficients randomly sampled from a uniform distribution within [-1,1] rad. This produces combined wavefront errors of approximately 2.0--3.5 rad, well beyond the 0.45 rad diffraction limit required for maximum theoretical resolution, representing severely aberrated conditions relevant to deep tissue microscopy.

\subsection{Phase-Diverse Image Generation.} 
Following previous work \cite{krishnan2020optical,kok2025practical}, we use phase-diverse images captured at multiple focal planes as input. The use of phase-diversity has been known to enhance models employing Zernikes by resolving the inherent ambiguity present in certain Zernike polynomials when looking at an image taken at a single focal plane. To generate our phase-diverse dataset, each sample image is convolved with an aberrated \ac{psf} at three defocus levels: negative defocus (-1 rad), at-focus (0), and positive defocus (+1 rad), as illustrated in Fig. \ref{fig:overall_pipeline}. For a typical microscope configuration (NA = 0.8, $\lambda$ = 800 nm), a defocus of $\pm1$ rad corresponds to a physical Z-axis displacement of approximately $\pm$0.4 $\mu$m.

\section{Methodology}
\label{sec:methodology}
In this section, we first delineate the overall architecture of \ac{zrnet}. Then we describe the proposed Zernike graph module and the Frequency-Aware Alignment (FAA) loss.

\subsection{Overall Pipeline}
\label{sec:pipeline}
As illustrated in Fig.~\ref{fig:overall_pipeline}, the proposed \ac{zrnet} processes phase-diverse aberrated images through two parallel branches: (1) An encoder-decoder architecture that employs KBNet$_S$ \cite{zhang2023kbnet} as its backbone for image restoration, and (2) a Zernike coefficient prediction branch featuring our Zernike Graph that enforces physical constraints by learning the established interactions between Zernike modes to decipher the nature of the aberration. The two branches mutually reinforce each other, compelling the image restoration to adhere to underlying optical properties while the restoration objective regularises Zernike coefficient estimation.

\subsection{Zernike Graph}
\label{sec:zernike_graph}
\begin{figure}[t!]
\begin{center}

   \includegraphics[width=1\linewidth]{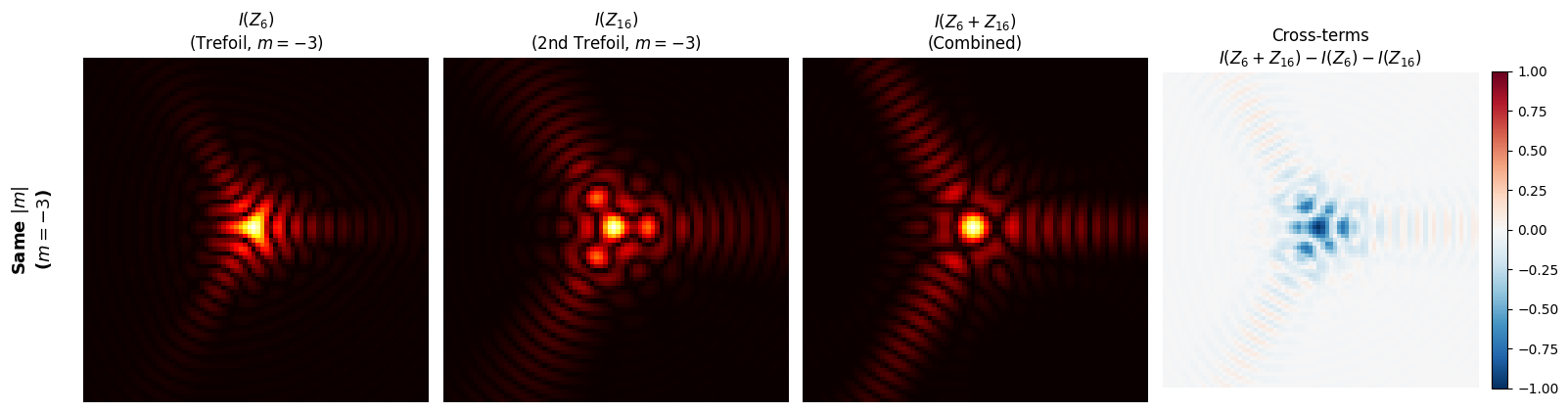}
\end{center}

   \caption{Demonstration of orthogonality breakdown in the intensity domain. The first two panels show the at-focus PSF intensity produced by individual Zernike modes $Z_6$ (vertical trefoil, $m=-3$) and $Z_{16}$ (vertical secondary trefoil, $m=-3$), each with coefficient amplitude 1.0 rad. The third panel shows the PSF when both modes are applied simultaneously. The fourth panel shows the cross-terms, computed as $I(Z_6 + Z_{16}) - I(Z_6) - I(Z_{16})$, which are clearly non-zero --- demonstrating that the combined intensity is not a simple sum of individual intensities.}
\label{fig:coupling}
\end{figure}

While Zernike polynomials are mathematically orthogonal when describing the optical field, the microscope measures intensity, which effectively breaks down this orthogonality. From Eqn.~\ref{eq:psf}, the microscope's intensity is computed by a squared modulus of the complex field. This squaring operation introduces cross-terms between pairs of Zernike modes, proportional to the product of their coefficients $c_j c_k$ from Eqn.~\ref{eq:phi}. In the intensity-domain, couplings dominate between modes sharing the same azimuthal degree (m), as these produce non-oscillatory (DC) components upon multiplication. This can be seen visually in the Zernike pyramid in Fig. \ref{fig:overall_pipeline} where Zernikes are grouped in vertical columns according to their angular frequency $m$. These observations directly motivate our graph construction (see Fig.~\ref{fig:coupling} for a visual demonstration of the coupling between two Zernike modes with m=-3 when looking in the intensity domain). We thus hypothesise that explicitly exploiting correlations among Zernike modes will improve aberration correction performance. 

However, standard deep learning approaches are not designed to capture these physical dynamics. They typically employ a fully connected layer to directly map the latent features extracted by the image encoder to the Zernike coefficients. This formulation effectively linearises the Zernike space into a flat, unstructured Euclidean vector, implicitly assuming that the modes are independent regression targets. Consequently, these models lack the structural capacity to explicitly model the conditional dependencies between coefficients, forcing the network to learn complex physical couplings entirely from data without architectural guidance.

In contrast, we argue that the non-linear coupling between Zernike modes is best modelled as a graph structure, where edges explicitly encode physical interactions based on Zernike mode groupings. To harness this topology, we leverage \ac{gnn}s, specifically \ac{gat} \cite{velickovic2017graph}, which excel at modelling inter-relationships through message passing between connected nodes. Inspired by QAGNet \cite{deng2024advancing}, we propose a grouped \ac{gnn} approach that moves beyond simple independent regression to reason about the aberration composition. Our approach achieves this by harnessing the dependencies both within natural Zernike groups and the global interactions between these groups. We consider two primary grouping strategies, and a baseline in which no grouping is applied:

\noindent\textbf{Azimuthal Degree Grouping.}
Zernike modes sharing the same azimuthal degree (m) have identical angular periodicity and symmetry. For example, vertical trefoil (Z6, m=-3) and vertical secondary trefoil (Z16, m=-3) collectively define a triple-angle sine variation, making them naturally suited for joint processing.

\noindent\textbf{Aberration Grouping.}
Organises Zernike polynomials based on the physical aberrations they represent in optical design. By grouping aberrations this way, Zernikes with the same radial order (n) and same absolute azimuthal degree (m) are assigned to the same class where their corresponding polynomials are identical with a $\pi/2$ phase shift. For instance, Z11 (n = 4, m=-2) and Z13 (n=4, m=2) together contribute to secondary astigmatism, creating a physically meaningful category for targeted aberration correction.

Our ablation studies indicate that organising Zernike polynomials by azimuthal degree yields superior results; hence, we focus on this strategy in our primary experiments. Further details regarding the aberration grouping and baseline are provided in \ref{appendix:zern_graph}.

As illustrated in Fig. \ref{fig:zernike_graphs}, the Zernike graph module operates in a hierarchical four stages process designed to model the physical relationships of aberrations:

\noindent\textbf{(1) Node Initialisation and Intra-Group Interaction.} First, the latent features extracted by the image encoder are projected into mode-specific representations via separate \ac{mlp}. Let $\mathcal{H}=\{h_3,…,h_{27}\}$ denote this set of latent feature vectors, which serve as the initialisation for the corresponding Zernike graph nodes. To capture local symmetry constraints, Zernike nodes are first partitioned into subsets $S_m$ based on their azimuthal degree $m \in M=\{-6,.. ,6\}$. For each subset, let $\mathbf{H}_{S_m}$ denote the feature matrix constructed by stacking the latent vectors $\{h_i \mid i \in S_m\}$. Within each group, a single \ac{gat} layer updates the features by modelling the intra-group dynamics. This deliberate choice of a single layer maintains a lean architecture that models the physical relationships of Zernike polynomials while achieving diffraction-limited performance:  

\begin{equation} 
\mathbf{\hat{H}}_{S_m} = \text{GAT}(\mathbf{H}_{S_m}, A_{intra})  \end{equation} where $\mathbf{\hat{H}}$ denotes the updated feature matrix and $A_{intra}$ represents the fully connected adjacency matrix within the group.

\noindent\textbf{(2) Group Aggregation via Proxy Nodes.} To capture the collective behaviour of each distinct angular variation, we aggregate the information within each subset $S_m$ into a unified proxy node $p_m$. This process consists of two steps. First, the proxy node is initialised as $p_m^0$ by averaging the original latent features $\mathbf{H}_{S_m}$:
\begin{equation} 
    p_m^0 = \text{Mean}(\mathbf{H}_{S_m}) 
\end{equation} 
Next, to refine this representation, we employ a directed message passing step where information flows from the updated constituent Zernike nodes ($\mathbf{\hat{H}}_{S_m}$) to the proxy:
\begin{equation} 
    p_m = \text{GAT}(p_m^0, \mathbf{\hat{H}}_{S_m} \mid A_{agg}) 
\end{equation} 
where $A_{agg}$ defines directed edges from the Zernike nodes to their respective proxy node. For singleton groups (e.g., Z15 for vertical pentafoil, $m=-5$), the node feature is directly assigned as the proxy.

\noindent\textbf{(3) Global Inter-Group Exchange.} To model the complex interaction between different angular symmetry patterns and how they influence each other within the optical system, the proxy nodes form a fully-connected graph. Let $\mathbf{P}$ denote the feature matrix constructed by stacking the proxy nodes $\{p_m \mid m \in M\}$. A \ac{gat} layer then facilitates the inter-group exchange:
\begin{equation} 
    \mathbf{\hat{P}} = \text{GAT}(\mathbf{P}, A_{global}) 
\end{equation} 
where $A_{global}$ represents the fully connected adjacency matrix.

\noindent\textbf{(4) Hierarchical Feedback and Prediction.} The processed group-level information is fed back to individual Zernike nodes to generate refined latent representations. To obtain the final coefficients, these refined features are passed through separate \ac{mlp}s to predict the final Zernike coefficients. This hierarchical approach ensures the framework respects both local relationships within azimuthal degree groups and global interactions between different angular symmetry variations, leading to more accurate and physically consistent predictions.

\begin{figure}[t!]
\begin{center}

   \includegraphics[width=1\linewidth]{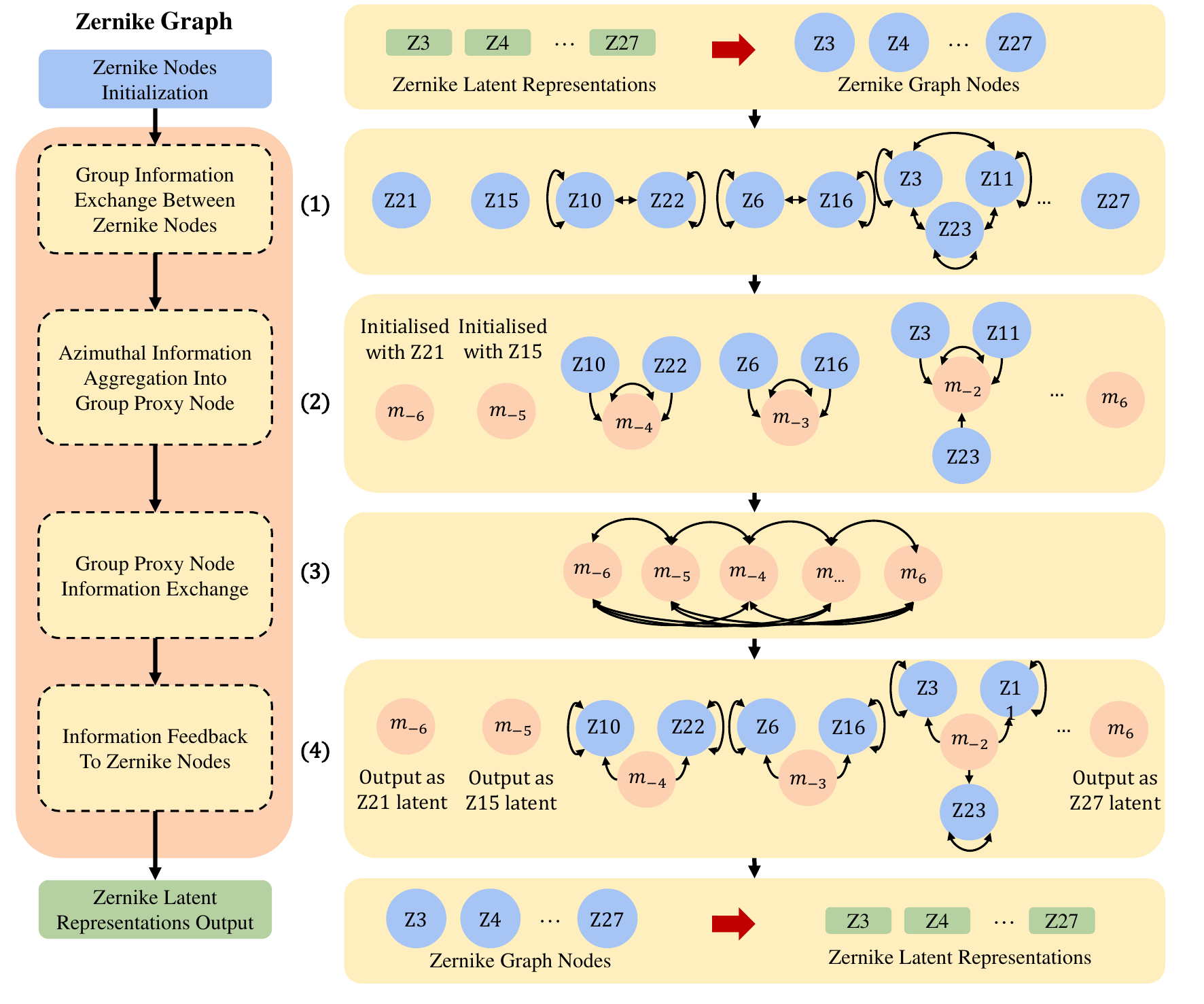}
\end{center}

   \caption{Our Zernike graph architecture for processing Zernike modes based on their azimuthal degrees. The framework processes information in four stages: (1) Node initialisation and intra-group exchange between Zernike nodes sharing the same azimuthal degree, (2) Information aggregation within each azimuthal degree group, (3) Inter-group information exchange via a fully-connected graph between different azimuthal degree groups, and (4) Information feedback to individual Zernike nodes to generate final latent representations.}
\label{fig:zernike_graphs}

\end{figure}

\subsection{Loss Function}
\label{sec:loss}

Traditional image restoration networks typically optimise an $L_1$ loss between the restored and ground truth images. Our framework extends this by introducing a parallel objective that minimises the \ac{mse} between predicted and ground truth Zernike coefficients --- fundamental optical descriptors that characterise wavefront aberrations. While this additional term helps guide the restoration process by leveraging aberration information encoded in the Zernike polynomials, optimising these losses independently can lead to physically inconsistent solutions. The network may learn shortcuts that, while effective on the training set, fail to capture the underlying optical physics and thus generalise poorly. 

To address this limitation, we introduce the \ac{faa} loss to enforce consistency between image restoration and Zernike prediction in the Fourier domain. This choice is motivated by our requirement to recover the high spatial frequencies that get lost in the image when aberrations are present. Different aberration types affect the frequency domain in distinct ways: for example, defocus produces isotropic attenuation of high spatial frequencies, while astigmatism creates an anisotropic frequency response, causing the image to become sharp in one direction and blurred in the other. Spatial-domain losses aggregate pixel-wise errors without distinguishing between these optically distinct degradation mechanisms, whereas operating in the Fourier domain exposes the frequency structure of these distortions, enabling more targeted penalisation. 

Our three-way \ac{faa} loss exploits the Fourier domain, capturing errors in both amplitude and phase structure. The loss ensures the network learns physically meaningful image restorative transformations, rather than merely minimising pixel-wise intensity errors, while simultaneously predicting Zernike coefficients that correspond to the actual restoration process, leading to improved generalisation and more reliable aberration correction. The \ac{faa} loss is formulated as:

\begin{equation}
L_{\text{FAA}} = L_{\text{R}} + L_{\text{C}} + L_{\text{Z}}
\end{equation}
where the three complementary terms in the equations enforce consistency in the frequency domain. Let $\mathscr{F}(x)$ denote the Fourier transform of $x$, $A(Z)$ represents the aberrated \ac{psf} generated from Zernike coefficients $Z$, $I_R$ and $I_{GT}$ be the restored and ground truth images respectively, and $Z_P$ and $Z_{GT}$ denote the predicted and ground truth Zernike coefficients. The operator $\circledast$ represents the convolution of the aberrated \ac{psf} with an image.

Each loss term computes the $L_1$ norm of both real ($Re$) and imaginary ($Im$) components separately in Fourier space, providing a numerically stable and computationally efficient way to capture frequency domain differences.
\begin{enumerate}[leftmargin=*]
  \item \textbf{Restoration Loss} \((L_{\text{R}})\):
The restoration loss ensures that the restored image, when re-aberrated with ground truth Zernikes, matches the original aberrated image in frequency space. 
\begin{equation}
    \begin{aligned}
        L_{\text{R}} = \sum_{p \in \{Re,Im\}} |p(&\mathscr{F}(I_R \circledast A(Z_{GT})) - \mathscr{F}(I_{GT} \circledast A(Z_{GT})))|
    \end{aligned}
\end{equation}

    \item  \textbf{Cross-verification Loss} \((L_{\text{C}})\):
Cross-verification loss bridges image restoration and Zernike prediction tasks by comparing the effects of ground truth Zernikes applied to the restored image against predicted Zernikes applied to the ground truth image, enforcing mutual consistency between the two tasks.
\begin{equation}
    \begin{aligned}
        L_{\text{C}} = \sum_{p \in \{Re,Im\}} |p(&\mathscr{F}(I_R \circledast A(Z_{GT})) - \mathscr{F}(I_{GT} \circledast A(Z_P)))|
    \end{aligned}
\end{equation}
    \item \textbf{Zernike Loss} \((L_{\text{Z}})\):
This loss assesses the accuracy of predicted Zernikes by comparing their effects with those of ground truth Zernikes when applied to the ground truth image.
\begin{equation}
   \begin{aligned}
       L_{\text{Z}} = \sum_{p \in \{Re,Im\}} |p(&\mathscr{F}(I_{GT} \circledast A(Z_{GT})) - \mathscr{F}(I_{GT} \circledast A(Z_P)))|
   \end{aligned}
\end{equation}
\end{enumerate}
 
The final total loss is defined as:
\begin{equation}
    \begin{aligned}
L_{\text{total}} = & L_1(I_{GT},I_{R}) + \lambda_1 MSE(Z_{GT},Z_{P}) + \lambda_2 L_{\text{FAA}}(I_{GT},I_{R},Z_{GT},Z_{P})
    \end{aligned}
    \label{eqn:loss}
\end{equation}
where $\lambda_1$ = 0.5 and $\lambda_2$ = 0.01 are weighting factors determined through grid search to balance the contributions of the loss terms.

\section{Experiments}
\label{sec:experiments}
\subsection{Experimental Setup}
\subsubsection{Dataset} 
\label{sub:data}
To demonstrate our model generalisation across diverse biological samples and microscopy modalities, we utilised CytoImageNet \cite{hua2021cytoimagenet}, a large-scale dataset designed for bioimage transfer learning, compiled from 40 openly available datasets within: Recursion, Image Data Resource (IDR) \cite{williams2017image}, Broad Bioimage Benchmark Collection (BBBC) \cite{ljosa2012annotated}, Kaggle, and Cell Image Library (CIL). The dataset comprises approximately 890k grayscale microscopy images across 894 classes, spanning fluorescence, confocal, brightfield, light, and darkfield modalities.

We train on the CytoImageNet training set of approximately 801k images. For efficient evaluation, we randomly select 20k images from the large validation set for our validation and another 20k images for testing. All sample images were resized to 256×256 pixels before convolution with aberrated PSFs to generate our phase-diverse dataset.

\subsubsection{Evaluation Metrics} 
For image restoration, we report standard image metrics PSNR, SSIM and LPIPS between the restored image and ground-truth images. To assess the accuracy of the Zernike predictions, we follow established method \cite{hu2023universal} in adaptive optics to use \ac{rmswfe} to quantify the residual wavefront error that would remain after correction by subtracting the predicted Zernike modes, $Z_{P}$ from the ground truth Zernike modes, $Z_{GT}$:
\begin{equation}\label{eq:rmswfe}
RMS_{WFE} = \sqrt{\sum_{j=1}^{M} \frac{1}{N} \sum_{i=1 }^{N} \left( Z_{{P}_{ij}} - Z_{{GT}_{ij}} \right)^2 } 
\end{equation}
where N is the number of samples and M is the number of Zernike modes. A \ac{rmswfe} $<$ 0.45 rad is generally considered indicative of diffraction-limited imaging performance \cite{marechal1947study}. To evaluate computational efficiency, we use calflops \cite{calflops} to calculate the floating-point operations (FLOPs) and report the mean inference time per image for each model.

\subsubsection{Implementation Details} 
We implement and train our models on phase-diverse aberrated images using a single NVIDIA RTX A6000 GPU with batch size 8. During each training iteration, Zernike coefficients are randomly sampled to generate aberrated \ac{psf} kernels, which are then convolved with optical images from the training dataset. Following KBNet$_S$ \cite{zhang2023kbnet} configuration, we train the model using fixed 256×256 images and the AdamW optimizer. We employ a cosine annealing scheduler with warm restarts. The learning rate is fixed at 3e-4 for the first 300k iterations, after which the scheduler begins warm restarts every 208k iterations with the learning rate decaying to a minimum of 2.5e-5.

Our training strategy consists of two phases: (1) We first pretrain the autoencoder and the initial \ac{mlp}s for 250k iterations, which excludes the Zernike graph and the final \ac{mlp} components. During this phase, the network learns direct Zernike coefficient prediction through the MLPs rather than generating latent representations for graph processing. This pretraining strategy is helpful as it allows the encoder-decoder backbone and coefficient prediction branches to establish robust feature representations prior to the introduction of more complex graph-structured learning. In contrast, training the full model from scratch without these prelearned features leads to unstable optimisation, where we observed loss divergence after approximately 60k iterations. During the pretraining stage, we exclude the \ac{faa} loss as initial Zernike predictions can be noisy and inconsistent, making any consistency constraints through the \ac{faa} loss counterproductive for training stability. (2) We then fine-tune the complete \ac{zrnet} framework for 1350k iterations, including the Zernike graph component, using identical training parameters and the complete loss function defined in Eq. \ref{eqn:loss}.

\subsection{Comparison with the State-Of-The-Art}
\label{sub:sota}
We compare the proposed \ac{zrnet} with several \ac{sota} image restoration and aberration correction networks: DeblurGANv2 \cite{kupyn2019deblurgan}, DeAbe (RCAN) \cite{guo2025deep}, SFT-DFCAN \cite{qiao2024deep}, MPRNet \cite{mehri2021mprnet}, Restormer \cite{zamir2022restormer}, CascadedGaze \cite{ghasemabadi2024cascadedgaze}, DiffIR \cite{xia2023diffir}, and KBNet (both large KBNet$_L$ and small KBNet$_S$ variants) \cite{zhang2023kbnet}. 

For Zernike coefficient prediction, we established a baseline using a variant of \ac{zrnet} without the image restoration component (\ac{zrnet} w/o image restoration). We also compared our framework against Swin Transformer~\cite{liu2021swin}.

To ensure fair comparison, we trained and evaluated all competing methods on the same dataset following their original configurations with minimal task-specific adaptations, and trained for identical durations as our approach. For each method, we selected the best-performing model based on validation set performance and report results from evaluating these models on our held-out test set. Detailed implementation specifications are provided in \ref{appendix:implementation}. 

\subsubsection{Quantitative Comparison} 
We present quantitative comparisons of image restoration in Tab. \ref{tb:quantitative_img}. Our method achieves the highest scores across all metrics, outperforming existing image restoration networks. Benefiting from physics-guided aberration modelling, our approach demonstrates superior pixel-wise accuracy (PSNR) and perceptual quality (SSIM, LPIPS), surpassing the second-best method KBNet$_L$ by 7.62\%, 6.27\%,  and 9.94\% respectively, while requiring 34.6\% fewer FLOPs and achieving 3.5$\times$ faster inference due to our use of a smaller 
backbone.

For Zernike coefficient prediction (Table \ref{tb:quantitative_zernike}), \ac{zrnet} reduces the initial aberration from an average \ac{rmswfe} of 2.8867 rad to 0.4374 rad, achieving diffraction-limited performance and outperforming other baselines. The joint training with image restoration substantially improves aberration characterisation, reducing the error by 18\% compared to the ablated variant (\ac{zrnet} w/o image restoration), highlighting the critical contribution of the image restoration branch for precise Zernike coefficient predictions. Even without image restoration, \ac{zrnet} outperforms Swin transformer, demonstrating the effectiveness of our \ac{gnn} architecture. 

Performance stratified by aberration severity 
(\ref{appendix:stratified}) shows that \ac{zrnet}'s advantage over KBNet$_L$ widens as aberration severity increases - the PSNR gap grows from 7.1\% (moderate) to 8.5\% (severe), with SSIM and LPIPS following the same trend — while Zernike predictions remain near diffraction-limited up to strong severity and exhibit only modest degradation under severe aberrations.

\begin{table}[t!]
\centering
\begin{tabular}{lccccc}
\toprule
Method & PSNR & SSIM & LPIPS & FLOPs (G) & Inference Time (ms) \\ 
\midrule
\multicolumn{6}{c}{\textit{General image restoration networks}} \\
DeblurGANv2 & 22.28 & 0.5659& 0.5187  & 51.06 & 23.17 \\
MPRNet & 23.47 & 0.6528 & 0.5088 & 1536.08 & 85.23 \\
Restormer & 23.62 & 0.6545 & 0.5027 & 282.01 & 108.25 \\
CascadedGaze & 24.59 & 0.6701& 0.4953  & 123.73 & 36.99 \\
DiffIR & 25.42 & 0.7039& 0.4610  & 225.41 & 92.98 \\ 
KBNet$_S$ & 26.22 & 0.7221 & 0.4512 & 137.62 & 100.00 \\ 
KBNet$_L$ & 26.76 & 0.7380 & 0.4316 & 215.72 & 361.22 \\ 
\multicolumn{6}{c}{\textit{Aberration correction networks}} \\
DeAbe (RCAN) & 21.30 & 0.5951& 0.5599  & 513.35 & 49.87 \\
SFT-DFCAN & 22.07 & 0.6113 & 0.5454 & 316.07 & 183.96 \\
ZRNet (Ours) & \textbf{28.80} & \textbf{0.7843}& \textbf{0.3887 } & 141.07 & 102.97 \\
\bottomrule
\end{tabular}
\caption{Quantitative results of image restoration on CytoImageNet \cite{hua2021cytoimagenet}. The images were convolved with optical aberrations with average \ac{rmswfe} of 2.8867 rad. Best results are highlighted in bold.}
\label{tb:quantitative_img}
\end{table}

\begin{table}[t!]
\centering
\begin{tabular}{lccc}
\toprule
Method & Zernike \ac{rmswfe} (rad) & FLOPs (G) & Inference Time (ms) \\ 
\midrule
Swin-B & 1.8846& 39.58 & 16.00 \\
Swin-L & 1.7576& 88.97 & 17.57 \\
ZRNet (w/o image restoration) & 0.5373& 66.22 & 53.58 \\
ZRNet & \textbf{0.4374}& 141.07 & 102.97\\
\bottomrule
\end{tabular}
\caption{Quantitative results of Zernike coefficient prediction on CytoImageNet \cite{hua2021cytoimagenet}. The images were convolved with optical aberrations with average \ac{rmswfe} of 2.8867 rad. Best results are highlighted in bold.}
\label{tb:quantitative_zernike}
\end{table}

\subsubsection{Qualitative Comparison} 
\begin{landscape}
    \begin{figure*}
        \centering
    	\setlength{\tabcolsep}{0.5pt}
        \renewcommand{\arraystretch}{0.5}
        	\begin{tabular}{ccccccccccc}

                \includegraphics[width=0.091\linewidth]{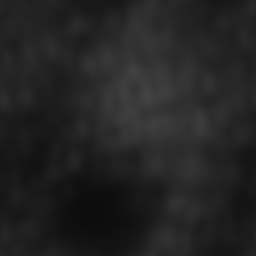} &
        		\includegraphics[width=0.091\linewidth]{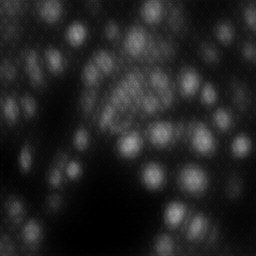} &
                 \includegraphics[width=0.091\linewidth]{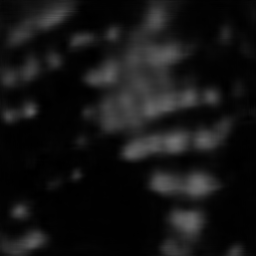} &
        		\includegraphics[width=0.091\linewidth]{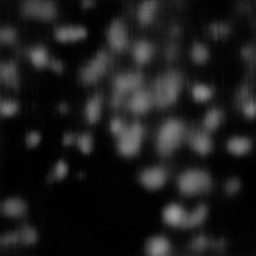} &
        		\includegraphics[width=0.091\linewidth]{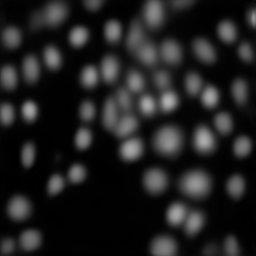} &
        		\includegraphics[width=0.091\linewidth]{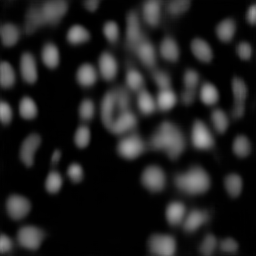} &
        		\includegraphics[width=0.091\linewidth]{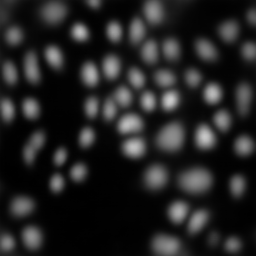} &
                \includegraphics[width=0.091\linewidth]{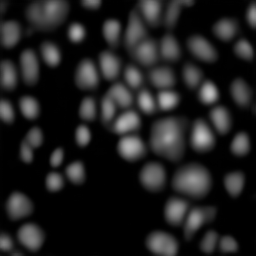} &
        		\includegraphics[width=0.091\linewidth]{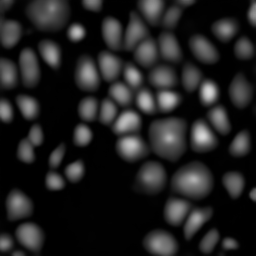} &
        		\includegraphics[width=0.091\linewidth]{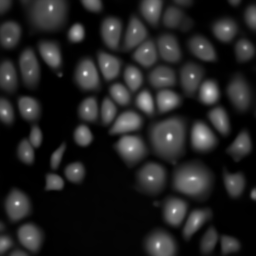} &
        		\includegraphics[width=0.091\linewidth]{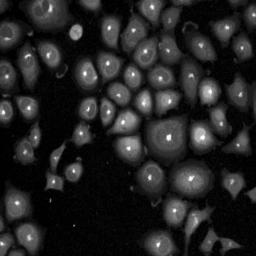} 
        		\\
        		\includegraphics[width=0.091\linewidth]{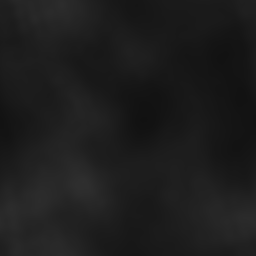} &
        		\includegraphics[width=0.091\linewidth]{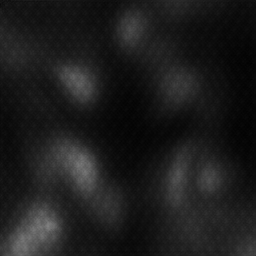} &
        		\includegraphics[width=0.091\linewidth]{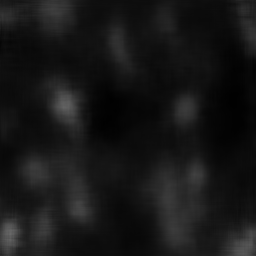} &
                	\includegraphics[width=0.091\linewidth]{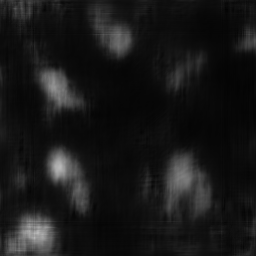} &
        		\includegraphics[width=0.091\linewidth]{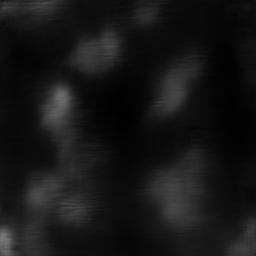} &
        		\includegraphics[width=0.091\linewidth]{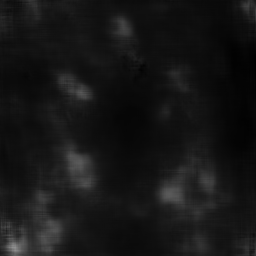} &
        		\includegraphics[width=0.091\linewidth]{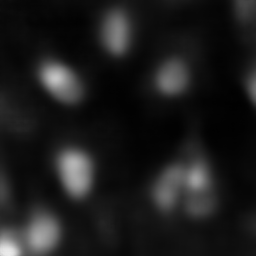} &
                \includegraphics[width=0.091\linewidth]{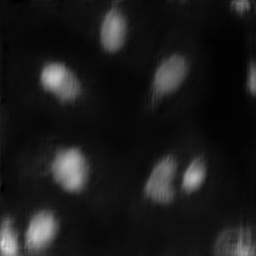} &
        		\includegraphics[width=0.091\linewidth]{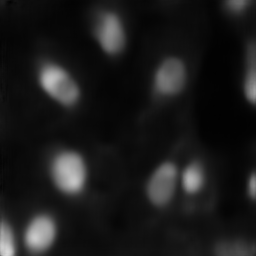} &
                        		\includegraphics[width=0.091\linewidth]{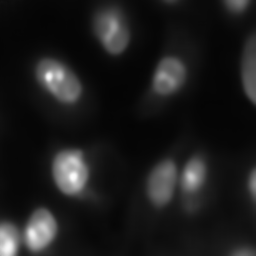} &
                                \includegraphics[width=0.091\linewidth]{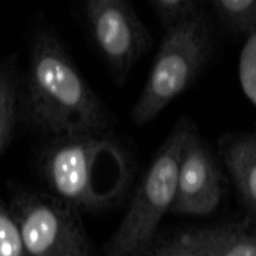} 

        		\\
 
                \includegraphics[width=0.091\linewidth]{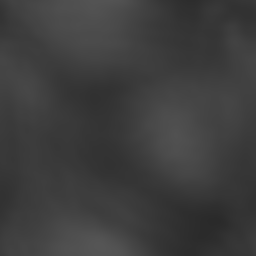} &
        		\includegraphics[width=0.091\linewidth]{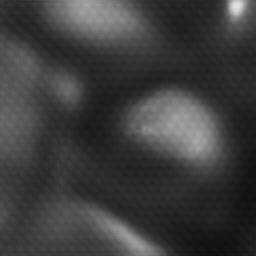} &
                 \includegraphics[width=0.091\linewidth]{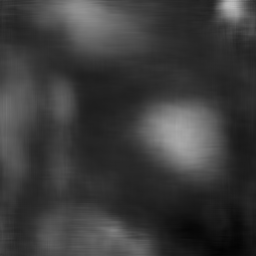} &
        		\includegraphics[width=0.091\linewidth]{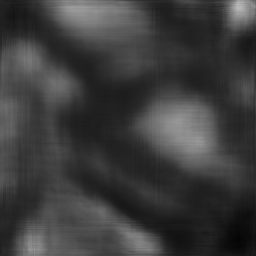} &
        		\includegraphics[width=0.091\linewidth]{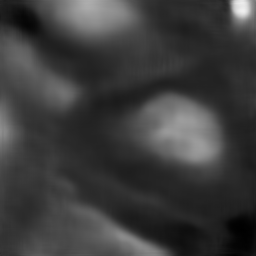} &
        		\includegraphics[width=0.091\linewidth]{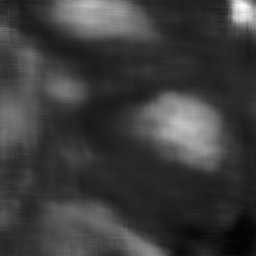} &
        		\includegraphics[width=0.091\linewidth]{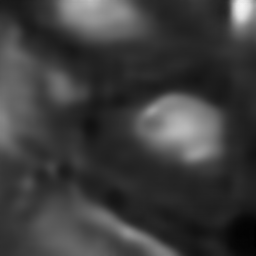} &
                \includegraphics[width=0.091\linewidth]{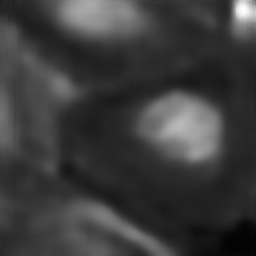} &
        		\includegraphics[width=0.091\linewidth]{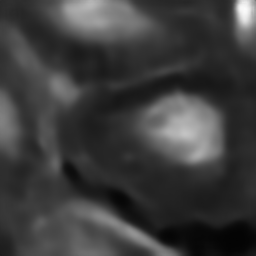} &
        		\includegraphics[width=0.091\linewidth]{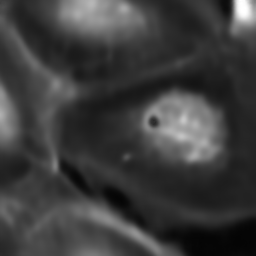} &
        		\includegraphics[width=0.091\linewidth]{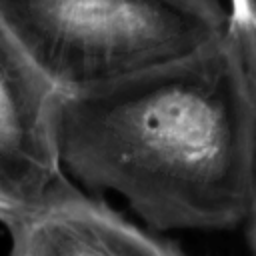} 
        		\\
                \includegraphics[width=0.091\linewidth]{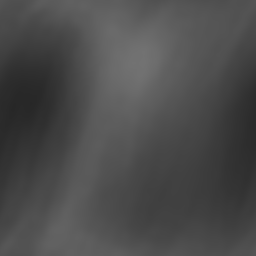} &
        		\includegraphics[width=0.091\linewidth]{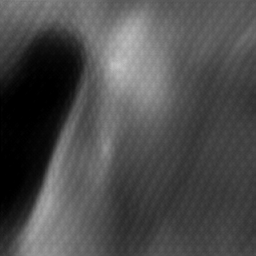} &
                 \includegraphics[width=0.091\linewidth]{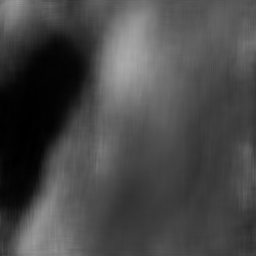} &
        		\includegraphics[width=0.091\linewidth]{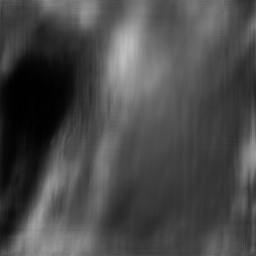} &
        		\includegraphics[width=0.091\linewidth]{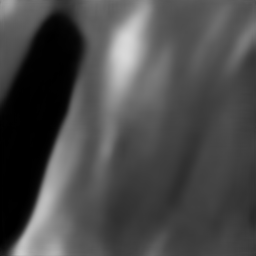} &
        		\includegraphics[width=0.091\linewidth]{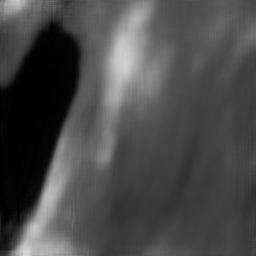} &
        		\includegraphics[width=0.091\linewidth]{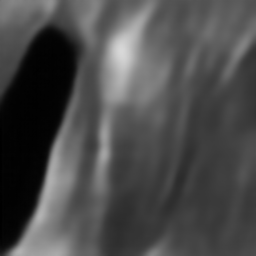} &
                \includegraphics[width=0.091\linewidth]{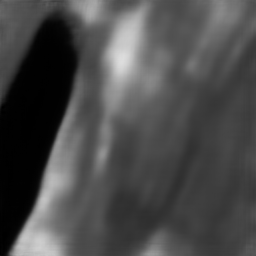} &
        		\includegraphics[width=0.091\linewidth]{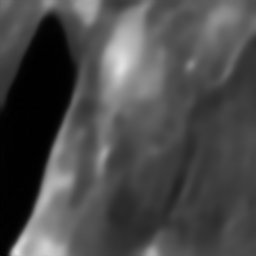} &
        		\includegraphics[width=0.091\linewidth]{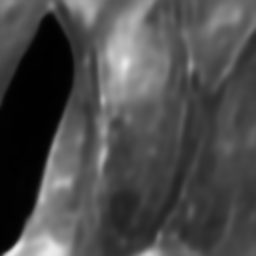} &
        		\includegraphics[width=0.091\linewidth]{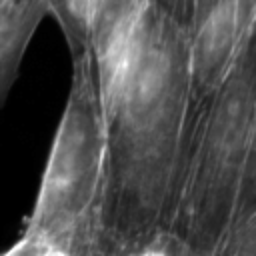} 
        		\\

                \includegraphics[width=0.091\linewidth]{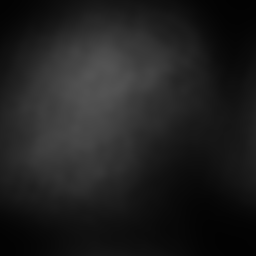} &
        		\includegraphics[width=0.091\linewidth]{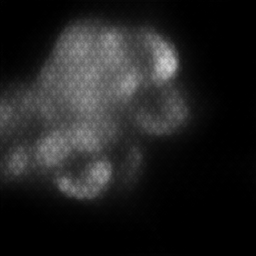} &
                 \includegraphics[width=0.091\linewidth]{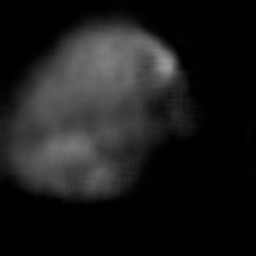} &
        		\includegraphics[width=0.091\linewidth]{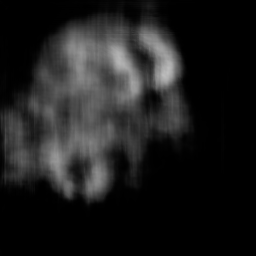} &
        		\includegraphics[width=0.091\linewidth]{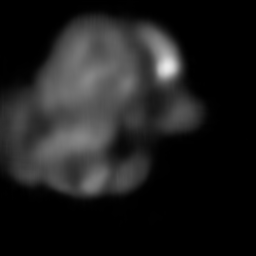} &
        		\includegraphics[width=0.091\linewidth]{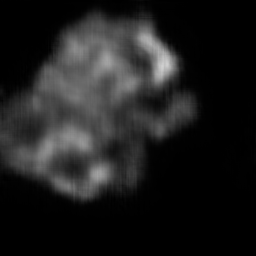} &
        		\includegraphics[width=0.091\linewidth]{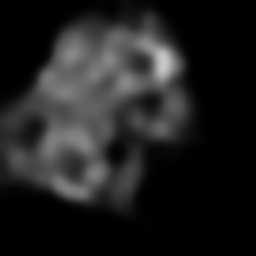} &
                \includegraphics[width=0.091\linewidth]{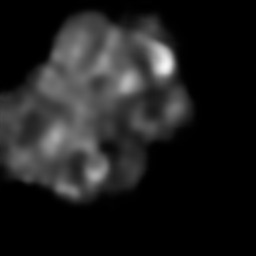} &
        		\includegraphics[width=0.091\linewidth]{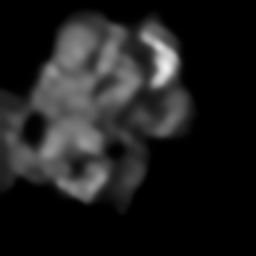} &
        		\includegraphics[width=0.091\linewidth]{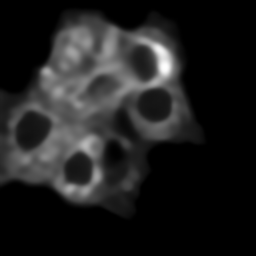} &
        		\includegraphics[width=0.091\linewidth]{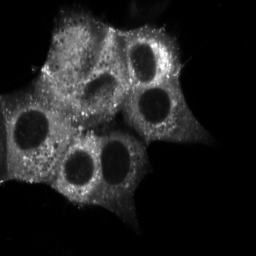} 
        		\\
                        
        		\includegraphics[width=0.091\linewidth]{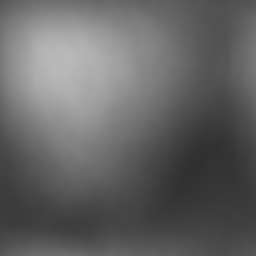} &
        		\includegraphics[width=0.091\linewidth]{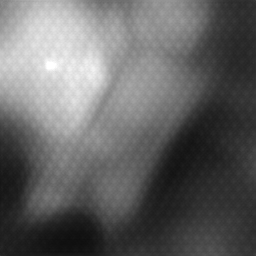} &
        		\includegraphics[width=0.091\linewidth]{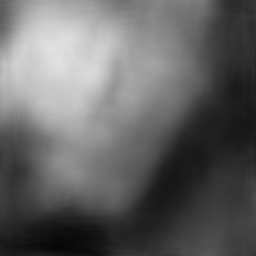} &
                	\includegraphics[width=0.091\linewidth]{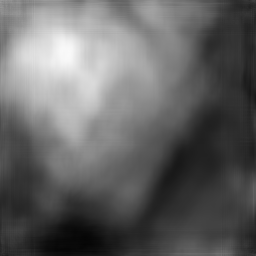} &
        		\includegraphics[width=0.091\linewidth]{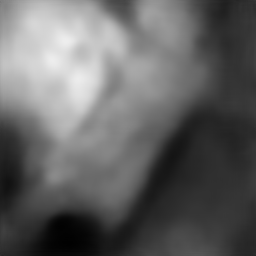} &
        		\includegraphics[width=0.091\linewidth]{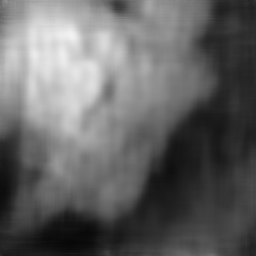} &
        		\includegraphics[width=0.091\linewidth]{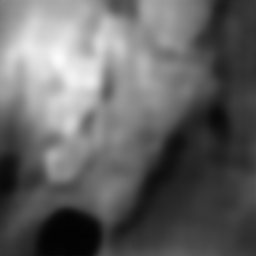} &
                \includegraphics[width=0.091\linewidth]{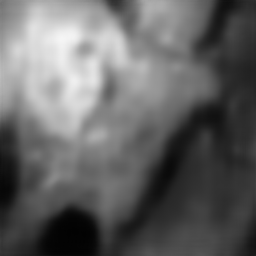} &
        		\includegraphics[width=0.091\linewidth]{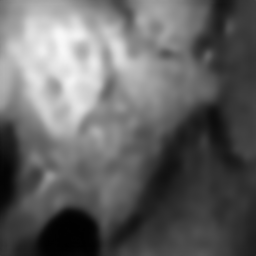} &
                        		\includegraphics[width=0.091\linewidth]{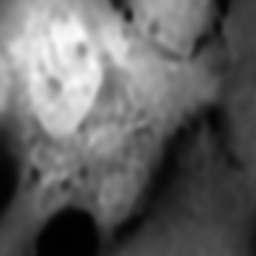} &
                                \includegraphics[width=0.091\linewidth]{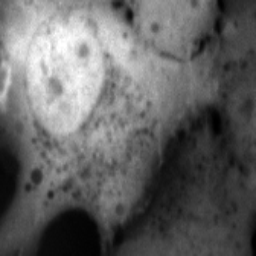} 

        		\\

        		\scriptsize{Input} & \scriptsize{DeblurGANv2} & \scriptsize{DeAbe (RCAN)} & \scriptsize{SFT-DFCAN}& \scriptsize{MPRNet} & \scriptsize{Restormer} & \scriptsize{CascadedGaze} & \scriptsize{DiffIR}  & \scriptsize{KBNet$_L$} & \scriptsize{\ac{zrnet}(Ours)} & \scriptsize{GT}  \\
        	\end{tabular}

	\caption{
	Qualitative comparison of \ac{sota} image restoration networks on CytoImageNet \cite{hua2021cytoimagenet} with optical aberration applied. ZRNet exhibits a higher level of fine detail after reconstruction, closer to the ground truth images.
	}
	\label{fig:qualitative_img}
	
\end{figure*}
\end{landscape}

\begin{figure*}[htbp]
    \begin{tikzpicture}
        \node[inner sep=0pt] (fig) at (0,0) {
            \includegraphics[width=0.8\textwidth]{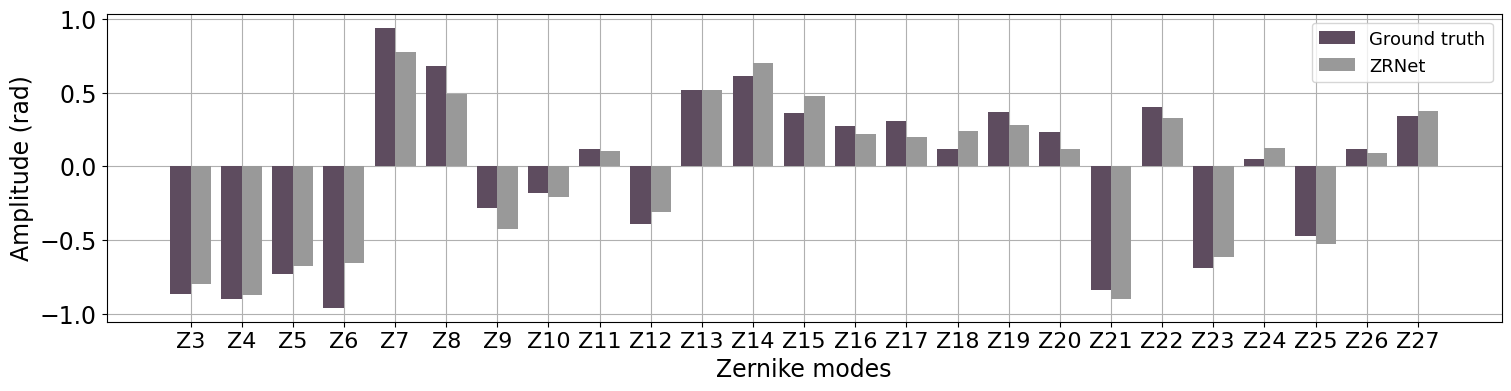}
        };
        \node[anchor=north west, font=\bfseries] at ($(fig.north west) + (-0.2,0.1)$) {(a)};
         \node[anchor=west, align=left, text width=4cm] at (5,0.2) {
            \textbf{Pre-correction \ac{rmswfe}}: \\2.7361 rad\\[0.2em]
            \textbf{After correction \ac{rmswfe}}\\ \ac{zrnet}: 0.5411 rad
        };
    
        \node[inner sep=0pt] (fig2) at (0,-3.5) {  
            \includegraphics[width=0.8\textwidth]{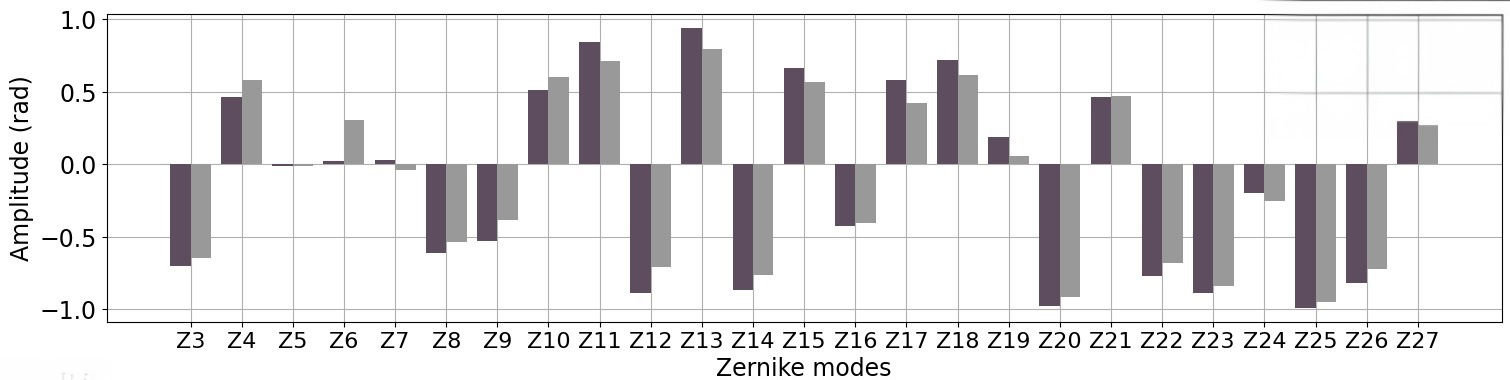}
        };
        
        \node[anchor=north west, font=\bfseries] at ($(fig2.north west) + (-0.2,0.1)$) {(b)};
        
        \node[anchor=west, align=left, text width=4cm] at (5,-3.2) {  
            \textbf{Pre-correction \ac{rmswfe}}:\\ 3.2570 rad\\[0.2em]
            \textbf{After correction \ac{rmswfe} }\\ \ac{zrnet}: 0.5576 rad
        };
        \node[inner sep=0pt] (fig3) at (0,-7) {  
            \includegraphics[width=0.8\textwidth]{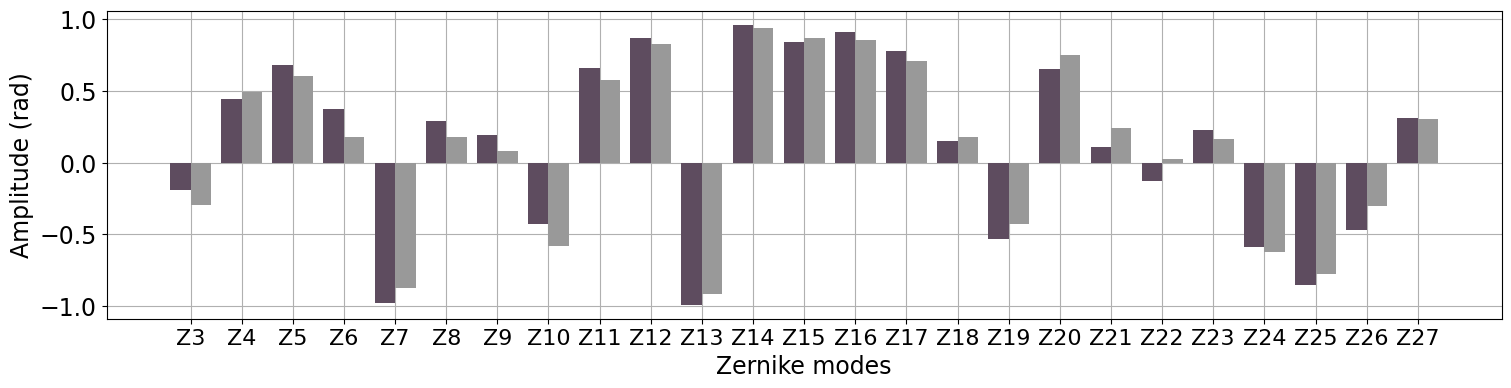}
        };
        
        \node[anchor=north west, font=\bfseries] at ($(fig3.north west) + (-0.2,0.1)$) {(c)};

        \node[anchor=west, align=left, text width=4cm] at (5,-6.8) {  
            \textbf{Pre-correction \ac{rmswfe}}: \\3.0865 rad\\[0.2em]
            \textbf{After correction \ac{rmswfe}}\\ \ac{zrnet}: 0.4901 rad
        };
  
        \node[inner sep=0pt] (fig4) at (0,-10.5) {  
            \includegraphics[width=0.8\textwidth]{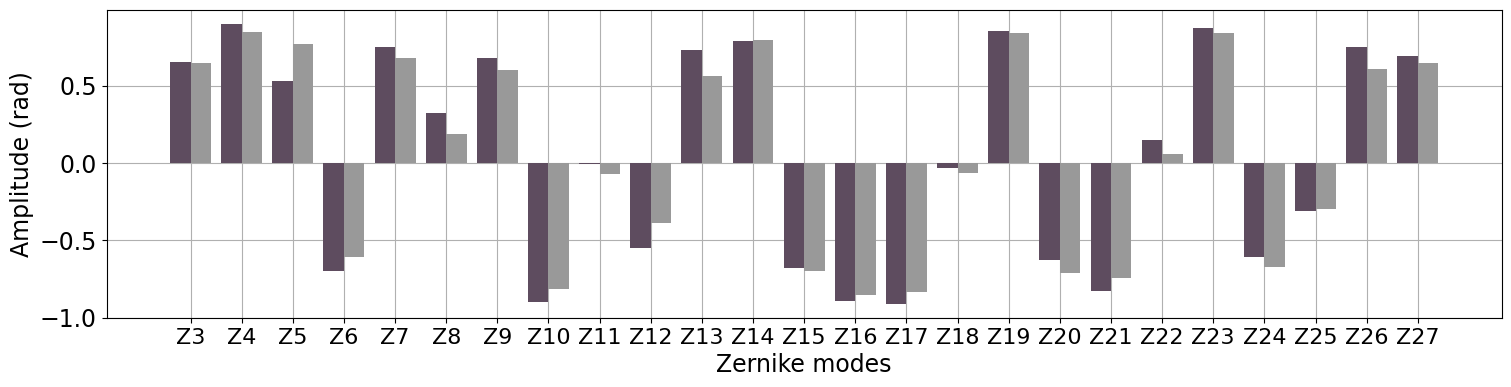}
        };

        \node[anchor=north west, font=\bfseries] at ($(fig4.north west) + (-0.2,0.1)$) {(d)};

        \node[anchor=west, align=left, text width=4cm] at (5,-10.3) {  
            \textbf{Pre-correction \ac{rmswfe}}: \\3.4027 rad\\[0.2em]
            \textbf{After correction \ac{rmswfe}}\\ \ac{zrnet}: 0.4698 rad
        };
        \node[inner sep=0pt] (fig5) at (0,-14) {  
            \includegraphics[width=0.8\textwidth]{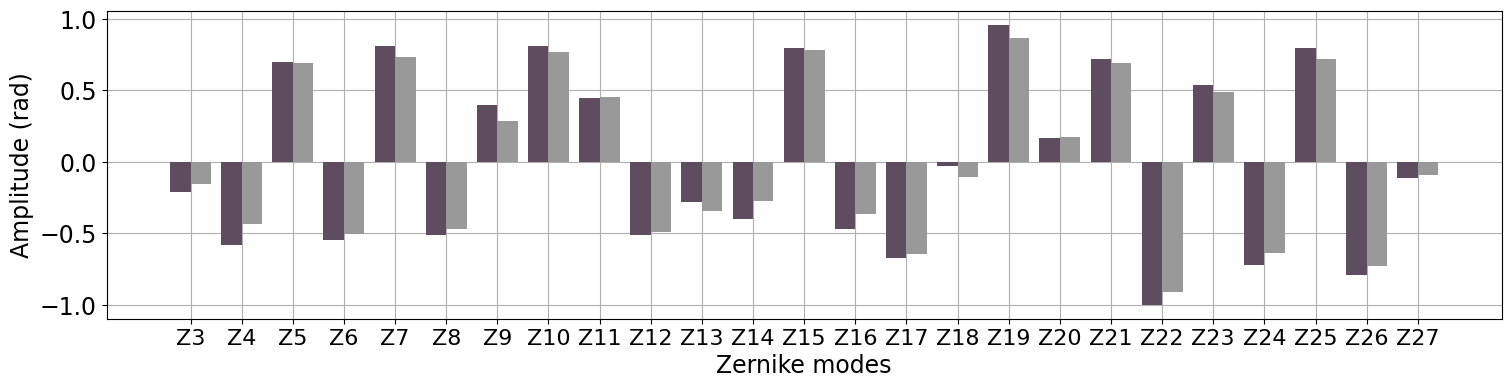}
        };
        
        \node[anchor=north west, font=\bfseries] at ($(fig5.north west) + (-0.2,0.1)$) {(e)};

        \node[anchor=west, align=left, text width=4cm] at (5,-13.8) {  
            \textbf{Pre-correction \ac{rmswfe}}:\\ 3.0777 rad\\[0.2em]
            \textbf{After correction \ac{rmswfe} }\\ \ac{zrnet}: 0.3511 rad
        };
        
        \node[inner sep=0pt] (fig6) at (0,-17.5) {  
            \includegraphics[width=0.8\textwidth]{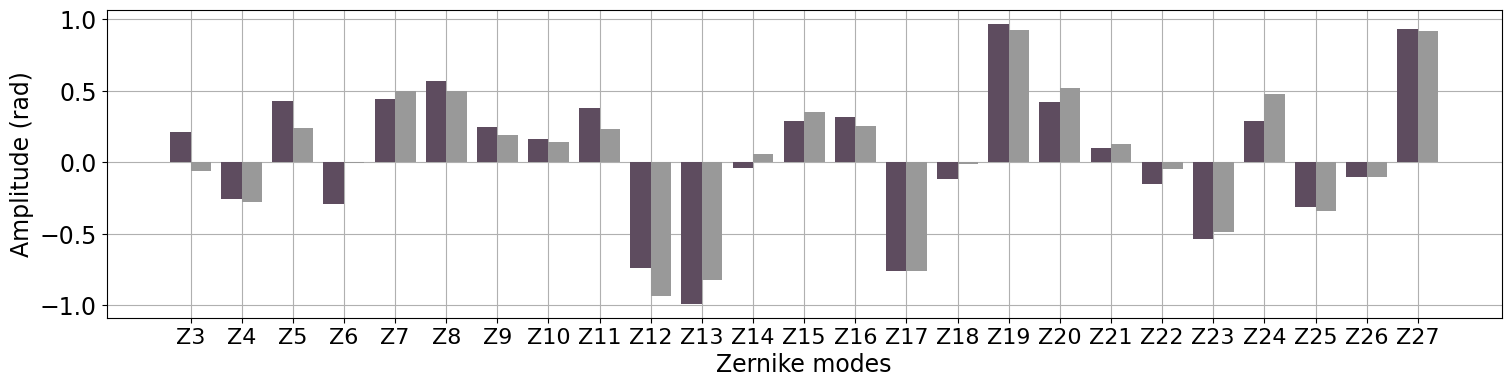}
        };
        
        \node[anchor=north west, font=\bfseries] at ($(fig6.north west) + (-0.2,0.1)$) {(f)};
        
        \node[anchor=west, align=left, text width=4cm] at (5,-17.3) {  
            \textbf{Pre-correction \ac{rmswfe}}:\\ 2.4364 rad\\[0.2em]
            \textbf{After correction \ac{rmswfe} }\\ \ac{zrnet}: 0.6199 rad
        };
    \end{tikzpicture}
    \caption{Zernike coefficients prediction comparing ground truth to ZRNet predictions.}
    \label{fig:qualitative_zernike}
\end{figure*}

Figure \ref{fig:qualitative_img} demonstrates qualitative comparisons with \ac{sota} methods on diverse cellular samples. Under severe optical aberrations, our method shows great improvement in restoration quality, preserving fine cellular details while other methods introduce artifacts or retain residual blur. Our restorations more closely match the ground truth, particularly in preserving cell morphology and intensity distributions. Corresponding Zernike coefficient predictions and pre-/post-correction \ac{psf}s are presented in Fig. \ref{fig:qualitative_zernike} and \ref{appendix:psf_comparison} respectively. In all cases, \ac{zrnet} accurately predicts both low and high-order Zernike modes, resulting in effective wavefront error reduction and consistent near diffraction-limited correction across diverse aberration patterns.

\begin{table*}[!t]
\centering
\scalebox{0.73}{
\begin{tabular}{lcccccccccc}
\toprule
\multirow{5}{*}{Set} & \multirow{5}{*}{\begin{tabular}[c]{@{}c@{}}Image\\restoration\\branch\end{tabular}} & \multirow{5}{*}{\begin{tabular}[c]{@{}c@{}}\ac{faa}\\loss\end{tabular}} & \multicolumn{4}{c}{Zernike coefficient prediction} & \multirow{5}{*}{PSNR} & \multirow{5}{*}{SSIM} & \multirow{5}{*}{LPIPS} & \multirow{5}{*}{\begin{tabular}[c]{@{}c@{}}Zernike\\\ac{rmswfe}\\(rad)\end{tabular}} \\
\cmidrule{4-7}& & &\multirow{3}{*}{\begin{tabular}[c]{@{}c@{}}MLP\\only\end{tabular}} & \multicolumn{3}{c}{Zernike graph} & & & \\\cmidrule{5-7} & && & No & Aberration & Azimuthal & & & \\& && & grouping & group & degree & & & & \\
\midrule
I & & & & & & \begin{tabular}[c]{@{}c@{}}\centering\checkmark\end{tabular} & NA & NA & NA&0.5373 \\
II & \begin{tabular}[c]{@{}c@{}}\centering\checkmark\end{tabular} & & & & & & 26.22 & 0.7221 &0.4512& NA \\
III & \begin{tabular}[c]{@{}c@{}}\centering\checkmark\end{tabular} & & \begin{tabular}[c]{@{}c@{}}\centering\checkmark\end{tabular} & & & & 27.70 & 0.7591 &0.4159& 1.6109 \\
IV & \begin{tabular}[c]{@{}c@{}}\centering\checkmark\end{tabular} &\begin{tabular}[c]{@{}c@{}}\centering\checkmark\end{tabular} & \begin{tabular}[c]{@{}c@{}}\centering\checkmark\end{tabular} & & & & 27.34 & 0.7500&0.4226 & 2.6799 \\ 
V & \begin{tabular}[c]{@{}c@{}}\centering\checkmark\end{tabular} & & & & & \begin{tabular}[c]{@{}c@{}}\centering\checkmark\end{tabular} & 28.48 & 0.7774&0.3962 & 0.4471 \\
VI & \begin{tabular}[c]{@{}c@{}}\centering\checkmark\end{tabular} & \begin{tabular}[c]{@{}c@{}}\centering\checkmark\end{tabular} & & \begin{tabular}[c]{@{}c@{}}\centering\checkmark\end{tabular} & & & 28.36 & 0.7744&0.3991 & 0.4652 \\
VII & \begin{tabular}[c]{@{}c@{}}\centering\checkmark\end{tabular} & \begin{tabular}[c]{@{}c@{}}\centering\checkmark\end{tabular} & & & \begin{tabular}[c]{@{}c@{}}\centering\checkmark\end{tabular} & & 28.60 & 0.7796 &0.3939& 0.4624 \\
VIII & \begin{tabular}[c]{@{}c@{}}\centering\checkmark\end{tabular} & \begin{tabular}[c]{@{}c@{}}\centering\checkmark\end{tabular} & & & & \begin{tabular}[c]{@{}c@{}}\centering\checkmark\end{tabular} & \textbf{28.80} & \textbf{0.7843}&\textbf{0.3887} & \textbf{0.4374} \\
\bottomrule
\end{tabular}
}
\caption{Ablation analysis of different modules in \ac{zrnet}.}
\label{tb:ablation}

\end{table*}

\subsection{Ablation Studies}
\label{sec:ablation_studies}
Table \ref{tb:ablation} presents our ablation analysis of the proposed designs.

\subsubsection{Joint Training Benefit} 
Removing the image restoration branch (Set I) hinders the network from learning how aberrations affect image quality, resulting in worse performance compared to Set V, thus confirming that image restoration provides essential supervision for accurate Zernike prediction through its physical constraints. Even with a suboptimal MLP only Zernike predictor (Set III and IV), joint training yields higher image quality metrics than restoration-only (Set II). This indicates that even imperfect Zernike coefficient predictions provide valuable guidance. Upgrading to our Zernike graph architecture (Sets V-VIII) further enhances restoration performance, demonstrating the mutual benefits between aberration estimation and image restoration.

\subsubsection{Effectiveness of \ac{faa} Loss} 
Applying the \ac{faa} loss to the MLP-only model (Set III vs IV) is counterproductive, as imposing a strict, physics-driven consistency constraint on the relatively noisy and naive MLP predictions creates a conflicting optimisation problem. In contrast, the full potential of \ac{faa} loss is realised when combined with Zernike graph (Set V vs VIII), which is explicitly designed to model relationships among Zernike modes and produce more coherent predictions, allowing \ac{faa} loss to function synergistically. Ablation studies for each of the FAA loss components, validating our three-way design, are provided in \ref{appendix:faa}.

\subsubsection{Importance of Zernike Graph} 
The addition of the Zernike graph in Set VIII leads to substantial improvements across all metrics, most notably in Zernike coefficient prediction when compared to Set IV. Crucially, this performance boost incurs negligible computational overhead, adding only 3.45 GFLOPs to the $KBNet_S$ backbone. This confirms that the structured Zernike graph better captures the relationship than the simple MLP with minimal additional complexity.

\subsubsection{Effect of Zernike Polynomial Groupings in Zernike Graph} 
In Sets VI to VIII, we include the \ac{faa} loss and analyse the different Zernike polynomial groupings in the Zernike graph. The results show that both physics-inspired grouping approaches (Sets VII and VIII) perform more effectively than no grouping (Set VI) across all image evaluation metrics, suggesting that structured Zernike relationships enhances restoration quality. Overall, grouping by azimuthal degree (Set VIII) achieves the best performance, indicating this approach better captures the optical principles governing wavefront behaviour and validating our hypothesis that modelling Zernike mode correlations is an effective strategy for aberration correction.

\subsection{Noise Robustness}
\label{sec:noise}
To evaluate robustness to realistic sensor noise, we test under three fixed noise conditions of increasing severity, each combining Poisson shot noise (parameterised by photon count $\alpha$) and Gaussian readout noise (standard deviation $\sigma$): mild ($\alpha = 1000$, $\sigma = 0.005$), moderate ($\alpha = 300$, $\sigma = 0.01$), and severe ($\alpha = 100$, $\sigma = 0.02$). All models are evaluated on the same 20k held-out test set used in Section~\ref{sub:data}. Training details are provided in \ref{appendix:noise}.

As shown in Tables~\ref{tb:noise_restoration} and~\ref{tb:noise_zernike}, models trained on clean data are not robust to noise — both \ac{zrnet} and KBNet$_L$ show substantial performance drops under noisy conditions, which is expected as neither was exposed to noise during training. To address this, we fine-tuned \ac{zrnet} and KBNet$_L$ from their clean-trained checkpoints on noise-augmented training data. The noise-augmented \ac{zrnet} maintains reasonable restoration quality even under severe noise and still reducing the average uncorrected \ac{rmswfe} from 2.8867 to 2.0116 rad (30.3\% reduction), while outperforming KBNet$_L$ across all noise conditions. The noise-augmented \ac{zrnet} shows a modest reduction on clean data compared to the clean-trained model, which is expected as the model learns to handle a broader range of input conditions.

\begin{table}[t!]
\centering
\begin{tabular}{llccc}
\toprule
Method & Noise Level & PSNR & SSIM & LPIPS \\
\midrule
\multirow{4}{*}{KBNetL (clean)} 
 & None   & 26.76  &0.7380  & 0.4316 \\
 & Mild   & 14.30 & 0.1370 &  0.7020\\
 & Moderate & 12.59 & 0.0783 &  0.7348\\
 & Severe & 11.689 & 0.0516 &0.7478  \\
\midrule
\multirow{4}{*}{\ac{zrnet} (clean)} 
 & None   & 28.80 & 0.7843 & 0.3887 \\
 & Mild   &  4.21 & 0.0023 & 0.8239 \\
 & Moderate &  4.19 & 0.0023 & 0.8237 \\
 & Severe &  4.23 & 0.0023 & 0.8237 \\
\midrule
\multirow{4}{*}{KBNetL (noise-aug)}
 & None   & 26.11 &0.7233  & 0.4518  \\
 & Mild   & 25.54  &0.7075  & 0.4692 \\
 & Moderate & 24.74 &0.6867  &0.4889  \\
 & Severe & 23.64 & 0.6578 & 0.5153  \\
\midrule
\multirow{4}{*}{\ac{zrnet} (noise-aug)} 
 & None   & 27.14 & 0.7499 & 0.4215 \\
 & Mild   & 26.44 & 0.7306 & 0.4471 \\
 & Moderate & 25.51 & 0.7067 & 0.4695 \\
 & Severe & 24.23 & 0.6728 & 0.5006 \\
\bottomrule
\end{tabular}
\caption{Noise robustness evaluation for image restoration.}
\label{tb:noise_restoration}
\end{table}

\begin{table}[t!]
\centering
\begin{tabular}{llc}
\toprule
Method & Noise Level & Zernike \ac{rmswfe} (rad) \\
\midrule
\multirow{4}{*}{\ac{zrnet} (clean)} 
 & None   & 0.4374 \\
 & Mild   & --- \\
 & Moderate & --- \\
 & Severe & --- \\
\midrule
\multirow{4}{*}{\ac{zrnet} (noise-aug)} 
 & None   & 1.2765 \\
 & Mild   & 1.4885 \\
 & Moderate & 1.7215 \\
 & Severe & 2.0116 \\
\bottomrule
\end{tabular}
\caption{Noise robustness evaluation for Zernike coefficient prediction. The input images were convolved with optical aberrations with average \ac{rmswfe} of 2.8867 rad. Results for the clean-trained model under noise are omitted (---) due to prediction failure, yielding physically meaningless errors ($>10^5$ rad).}
\label{tb:noise_zernike}
\end{table}

\begin{table}[t!]
\centering
\begin{tabular}{lc}
\toprule
Method & Zernike \ac{rmswfe} \\ 
\midrule
Swin-L & 2.6508 \\
\ac{zrnet} (w/o image restoration, MLP-only)& 1.6296 \\
\ac{zrnet} (w/o image restoration) & \textbf{1.6047} \\
\bottomrule
\end{tabular}
\caption{Zernike coefficient prediction on experimental PSF data~\cite{kok2025practical}. All models predict 24 Zernike coefficients (Z3, Z5--Z27) from phase-diverse PSF measurements. The images contain aberrations with average \ac{rmswfe} of 2.8203 rad. Best results are highlighted in bold.}
\label{tb:experimental_zernike}
\end{table}

\subsection{Experimental Validation on Real Aberrations}
\label{sec:real_psf}
To validate generalisation to real-world conditions, we evaluate Zernike prediction on experimental PSF data from~\cite{kok2025practical}, where a spatial light modulator was used to introduce known aberrations on a physical microscope, providing ground truth Zernike coefficients. The dataset comprises 15,000 phase-diverse PSF measurements across 24 Zernike modes (Z3, Z5--Z27, with Z4 excluded). Since the data consists of point source measurements, we evaluate the Zernike prediction branch independently without the image restoration component. Training details are provided in \ref{appendix:exp_psf}.

Results in Table~\ref{tb:experimental_zernike} show that \ac{zrnet} (w/o image restoration) achieves the lowest \ac{rmswfe}, reducing the average uncorrected \ac{rmswfe} by 43.1\% and substantially outperforming Swin-L. The MLP-only variant further confirms that the Zernike graph provides a consistent improvement even on experimental data. The prediction errors are higher than on synthetic data, which we attribute to the domain gap between simulated and physically measured aberrations, including non-Zernike distortions from the optical system not captured by the polynomial model. Nevertheless, sample predictions in Fig.~\ref{fig:real_zernike} show that the model correctly recovers the sign and approximate magnitude of individual Zernike coefficients across a range of aberration severities, with the wavefront error consistently reduced from the initial aberration.

\section{Conclusion}
\label{sec:conclusion}
In this work, we proposed \ac{zrnet}, a physics-informed framework that enables reliable optical image restoration with simultaneous Zernike coefficient prediction. We introduced a GNN-based Zernike graph architecture to model physical relationships between Zernike modes based on their azimuthal degrees, along with the \ac{faa} loss to enforce consistency between image reconstruction and aberration prediction in the frequency domain. Extensive experiments demonstrate that our approach achieves \ac{sota} performance across diverse microscopy modalities and biological samples in CytoImageNet, particularly in correcting high-order, large-amplitude aberrations where standard image restoration methods typically fail. These findings are further supported by experimental validation on real PSF data from a physical microscope and noise robustness experiments demonstrating robust performance under realistic sensor noise conditions. We have also shown, for the first time, that optimising the joint tasks of image restoration and Zernike prediction improves the performance of both sub-tasks, establishing new directions for integrating physical principles with deep learning for optical imaging applications.

While our framework shows robust performance on real microscopy images with synthetically applied aberrations, the domain gap between mathematically simulated aberrations and the complex, heterogeneous distortions found in physical optical systems remains a challenge. Additionally, our current model assumes spatially invariant aberrations, which is generally valid when working within the field of view specifications of the microscope objective. For spatially varying cases, which might occur under extreme aberration and imaging deep into a sample, a tiling approach with overlapping patches would be required but remains feasible given the fast inference time.

Future work will focus on bridging the gap between simulation and reality to enable practical deployment on experimental setups. Given the scarcity of large-scale paired experimental datasets, we plan to investigate data-efficient domain adaptation strategies. Specifically, exploring parameter-efficient fine-tuning and leveraging high-quality generative priors could allow the restoration model to adapt to specific experimental setups using minimal real-world samples. Furthermore, generative models such as diffusion models conditioned on optical parameters could be used to construct more realistic synthetic training data to reduce reliance on large-scale paired experimental datasets. To eventually transition these models to real-time edge environments, future work will also investigate efficiency improvements through knowledge distillation, model pruning, or reduced-precision inference. By extending the framework to handle these practical constraints, we aim to establish a generalisable, real-time aberration correction framework that expands the role of deep learning in cost-effective, high-resolution medical imaging.

\section*{CRediT authorship contribution statement}
\textbf{Yong En Kok:} Conceptualization, Data curation, Formal analysis, Investigation, Methodology, Project administration, Software, Validation, Visualization, Writing – original draft, Writing – review and editing.
\textbf{Bowen Deng:} Formal analysis, Investigation, Methodology, Visualization, Writing – review and editing.
\textbf{Alexander Bentley:} Formal analysis, Investigation, Writing – review and editing.
\textbf{Andrew J. Parkes:} Formal analysis, Supervision, Writing – review and editing.
\textbf{Michael G. Somekh:} Formal analysis, Supervision, Software, Writing – review and editing.
\textbf{Amanda J. Wright:} Formal analysis, Supervision, Writing – review and editing.
\textbf{Michael P. Pound:} Conceptualization, Formal analysis, Project administration, Supervision, Writing – review and editing.

\section{Funding}
This work was supported by the Engineering and Physical Sciences Research Council (EP/T020997/1) and University of Nottingham PhD studentship.

\section*{Declaration of competing interest}
The authors declare that they have no known competing financial interests or personal relationships that could have appeared to influence the work reported in this paper.


\appendix

\section{Zernike graph}
\label{appendix:zern_graph}
We evaluate three distinct strategies for organising Zernike graph: aberration grouping, azimuthal degree grouping, and a baseline with no grouping strategy. While the azimuthal degree grouping is detailed in the methodology section, here we discuss the aberration grouping approach and the baseline with no grouping.

\subsection{Aberration group}
Similar to Fig. 3 in the main paper, this framework follows the same four-stage architecture described in the methodology section but with aberration-based groups replacing azimuthal degree groups. By structuring the graph around aberration groups, the network explicitly learns how specific combinations of Zernike modes with similar optical physical effects collectively distort the wavefront and thereby facilitates effective aberration correction.

\subsection{Baseline with no grouping strategy}
In this baseline method, no grouping is applied; each node is initialised with its corresponding Zernike latent representation and exchanges information with all other Zernike modes through a fully connected graph structure. This fully connected approach does not explicitly model any physical relationships between the Zernike modes, providing a reference point for assessing the benefits of structured grouping strategies.

\section{Implementation details of competing \ac{sota} methods}
\label{appendix:implementation}
We implemented DeblurGANv2 \cite{kupyn2019deblurgan}, MPRNet \cite{mehri2021mprnet}, Restormer \cite{zamir2022restormer}, CascadedGaze \cite{ghasemabadi2024cascadedgaze}, DiffIR \cite{xia2023diffir}, and KBNet (both large KBNet$_L$ and small KBNet$_S$ variants) \cite{zhang2023kbnet} according to their official configurations. All these networks were originally designed for general image restoration tasks (e.g., denoising, deraining, deblurring); in our experiments, we adapted them to produce a single-channel restored optical image from the phase-diverse inputs and adjusted hyperparameters only when necessary to accommodate the larger dataset size and extended training duration. 

For specialised aberration correction networks, we compared against DeAbe (RCAN) \cite{guo2025deep} and SFT-DFCAN \cite{qiao2024deep}. The DeAbe model employs a 3D Residual Channel Attention Networks (RCAN) to remove aberrations from 3D fluorescence microscopy stacks and then uses two additional networks to progressively improve image resolution and contrast. This 3D RCAN is adapted from the 2D RCAN, which was originally developed for image super-resolution on natural images. In our work, we used only the 2D RCAN variant. Following insights from the DeAbe paper that microscopy images often have lower high-frequency content compared to natural images, we modified the network to process whole 256×256 phase-diverse images rather than patches, and adjusted the architecture to 5 residual groups and 10 residual blocks to accommodate the larger input size. The final layer of the network was further modified to output single-channel restored images.

For SFT-DFCAN \cite{qiao2024deep}, the original implementation requires first training SFE-Net to predict the \ac{psf} from aberrated images, then using both the aberrated image and estimated \ac{psf} as inputs to SFT-DFCAN to generate the super-resolved, aberration-corrected result. Given the complexity of our dataset and the authors' demonstration that accurate \ac{psf} estimation is critical for successful microscopy image reconstruction, we modified our implementation to  directly accept the ground truth 33×33 \ac{psf} together with the 256×256 phase-diverse input and then output the 256×256 restored single-channel image. This ensures a fair comparison by isolating the image restoration performance from potential errors in \ac{psf} estimation.

For networks that predict Zernike coefficients, we trained Swin Transformer (Base and Large variants) \cite{liu2021swin}  following standard protocols. We adapted each backbone to accept the same 256×256 phase-diverse inputs and modified the final layers to regress 25 Zernike coefficients.
\begin{table}[t!]
\centering
\begin{tabular}{ccccccc}
\toprule
\multirow{2}{*}{L$_R$}& \multirow{2}{*}{L$_C$}&\multirow{2}{*}{L$_Z$}&\multirow{2}{*}{PSNR}& \multirow{2}{*}{SSIM}& \multirow{2}{*}{LPIPS}& Zernike \\
&&&& & & RMS$_{WFE}$ (rad) \\

\midrule
\checkmark &&& 28.56& 0.7795 &0.3930 &0.4542\\
 &\checkmark&& 28.55& 0.7789 &0.3947 &0.4552\\
&&\checkmark& 28.53& 0.7779 &0.3945 &0.4551\\ 
\checkmark &\checkmark&& 28.57& 0.7785 &0.3938&0.4574\\
 &\checkmark&\checkmark& 28.54& 0.7778 &0.3947&0.4550\\
 \checkmark &&\checkmark& 28.56& 0.7794 &0.3930&0.4506\\
\checkmark &\checkmark&\checkmark&\textbf{28.80}& \textbf{0.7843} &\textbf{0.3887}&\textbf{0.4374}\\
\bottomrule
\end{tabular}

\caption{Ablation analysis for FAA loss.}
\label{tb:ab_faa}

\end{table}

\section{Performance Stratified by Aberration Severity}
\label{appendix:stratified}
Table~\ref{tb:stratified} presents performance stratified across three aberration severity bins: moderate (\ac{rmswfe} $<$ 2.8 rad), strong (2.8--3.0 rad), and severe ($>$ 3.0 rad). \ac{zrnet}'s advantage over KBNet$_L$ widens consistently with severity: the PSNR gap grows from 7.1\% to 8.5\%, SSIM from 5.5\% to 7.4\%, and LPIPS from 9.7\% to 10.7\% across moderate to severe bins. Zernike prediction accuracy remains near diffraction-limited through strong severity, with only modest degradation under severe aberrations.

\begin{table}[t!]
\centering
\begin{tabular}{llcccc}
\toprule
Severity (count) & Method & PSNR & SSIM & LPIPS & Pred \ac{rmswfe} \\
\midrule
\multirow{2}{*}{Moderate (7585)} 
 & KBNet$_L$ & 27.51 & 0.7571 & 0.4177 & --- \\
 & ZRNet     & \textbf{29.46} & \textbf{0.7986 }& \textbf{0.3771} & \textbf{0.3903} \\
\midrule
\multirow{2}{*}{Strong (5897)} 
 & KBNet$_L$ & 26.75 & 0.7373 & 0.4322 & --- \\
 & ZRNet     & \textbf{28.80} & \textbf{0.7840} & \textbf{0.3884} & \textbf{0.4415} \\
\midrule
\multirow{2}{*}{Severe (6518)} 
 & KBNet$_L$ & 25.87 & 0.7162 & 0.4476 & --- \\
 & ZRNet     & \textbf{28.07} & \textbf{0.7695 }& \textbf{0.3996} & \textbf{0.5050 }\\
\bottomrule
\end{tabular}
\caption{Performance stratified by aberration severity: moderate (\ac{rmswfe} $<$ 2.8 rad), strong (2.8--3.0 rad), and severe ($>$ 3.0 rad).}
\label{tb:stratified}
\end{table}

\section{PSF Comparison}
\label{appendix:psf_comparison}
\begin{figure*}[htbp]
\centering
\includegraphics[width=0.8\textwidth]{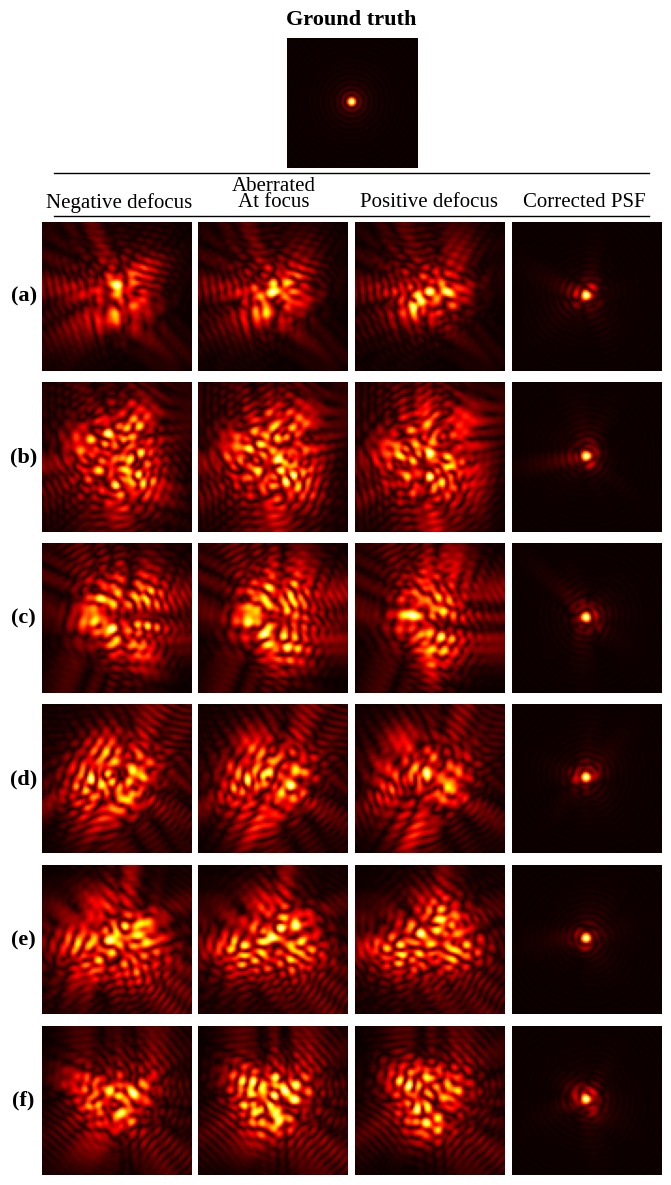}
\caption{Visualisation of aberration correction corresponding to the Zernike coefficient predictions in Fig.~\ref{fig:qualitative_zernike}. The ground truth (top) shows the diffraction-limited PSF. For each sample (a--f), the first three columns show the phase-diverse aberrated PSFs used to generate the degraded input images. The fourth column shows the PSF reconstructed from the residual Zernike coefficients (ground truth minus predicted), representing the aberration that would remain after applying ZRNet's predicted correction. }
\label{fig:psf_comparison}
\end{figure*}

Figure \ref{fig:psf_comparison} shows the visualisation of aberration correction corresponding to the Zernike coefficient predictions in Fig.~\ref{fig:qualitative_zernike}. 

\section{Ablation studies on \ac{faa} loss components}
\label{appendix:faa}
Table~\ref{tb:ab_faa} presents an ablation study evaluating the individual and combined effects of the three FAA loss components: Restoration loss $(L_{\text{R}})$, Cross-verification loss $(L_{\text{C}})$, and Zernike loss $(L_{\text{Z}})$. While individual components or pairs of two offer marginal gains, some combinations are counterproductive. The best performance is achieved when all three components are used together, confirming that the full FAA loss provides complementary, three-way supervision that effectively aligns image restoration with physically consistent Zernike prediction in the spatial frequency domain.

\section{Noise Robustness Training Details}
\label{appendix:noise}
For the noise robustness experiments, both ZRNet and KBNet$_L$ were fine-tuned from their respective clean-trained checkpoints on 10k images randomly selected from the CytoImageNet training set, with 1k images for validation. During training, combined Poisson shot noise and Gaussian readout noise were applied on-the-fly with parameters sampled uniformly across the full noise range: Poisson $\alpha$ was sampled log-uniformly from $[100, 1000]$ and Gaussian $\sigma$ was sampled uniformly from $[0.005, 0.02]$, ensuring equal representation of mild through severe noise conditions. Validation used a fixed moderate noise level ($\alpha = 300$, $\sigma = 0.01$). Both models were fine-tuned for 100 epochs using the AdamW optimiser with learning rate 2e-4, weight decay 5e-5, and cosine annealing (minimum learning rate 1e-5) with 15-epoch warmup. KBNet$_L$ was trained with $L_1$ pixel loss only, while ZRNet was trained using the complete loss function defined in Eq.~\ref{eqn:loss}. 

\section{Experimental PSF Training Details and Results}
\label{appendix:exp_psf}
For the experimental \ac{psf} evaluation, the dataset was split into 12,000 training, 1,500 validation, and 1,500 test samples. Swin-L and both ZRNet variants were trained from scratch using the KBNetS encoder with the Zernike prediction branch only, optimising MSE loss on 24 Zernike coefficients (Z3, Z5--Z27). Training used the AdamW optimiser with learning rate 5e-4, weight decay 5e-2, gradient clipping at 1.0, and cosine annealing with a 20-epoch warmup, for 100 epochs. 

\begin{figure*}[htbp]
    \begin{tikzpicture}
        \node[inner sep=0pt] (fig) at (0,0) {
            \includegraphics[width=0.8\textwidth]{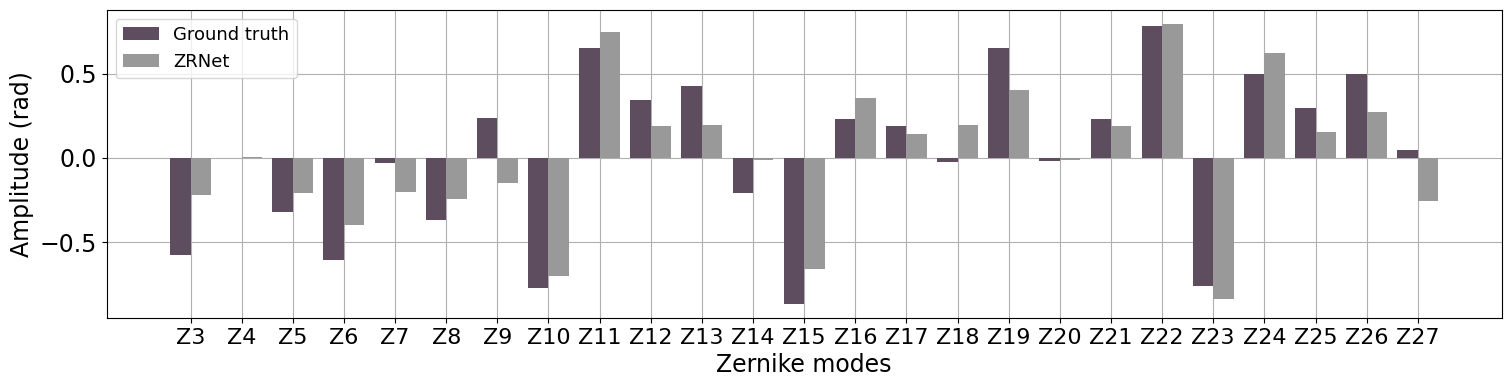}
        };
        \node[anchor=north west, font=\bfseries] at ($(fig.north west) + (-0.2,0.1)$) {(a)};
         \node[anchor=west, align=left, text width=4cm] at (5,0.2) {
            \textbf{Pre-correction \ac{rmswfe}}: \\2.3379 rad\\[0.2em]
            \textbf{After correction \ac{rmswfe}}\\ \ac{zrnet}: 0.9339 rad
        };
    
        \node[inner sep=0pt] (fig2) at (0,-3.5) {  
            \includegraphics[width=0.8\textwidth]{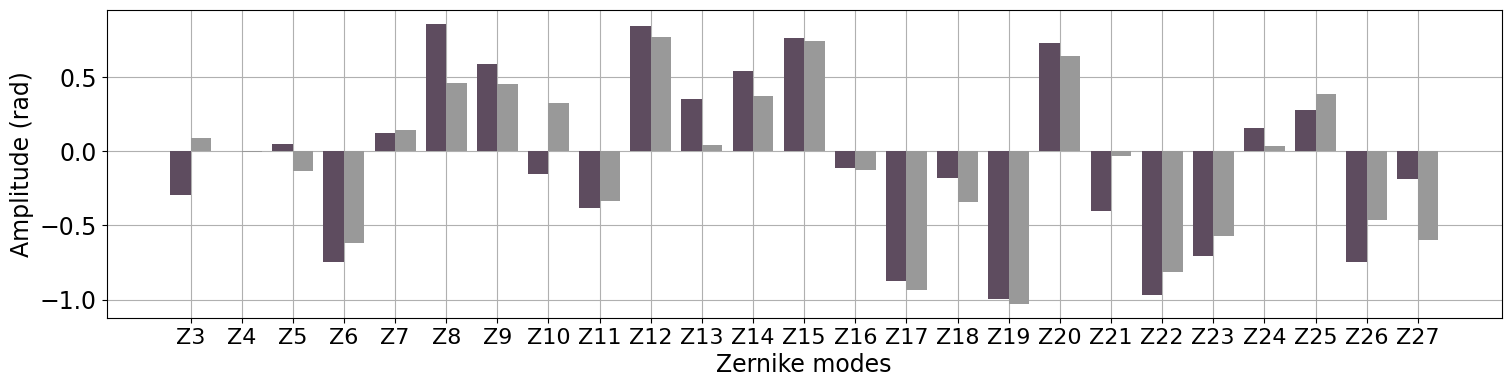}
        };
        
        \node[anchor=north west, font=\bfseries] at ($(fig2.north west) + (-0.2,0.1)$) {(b)};
        
        \node[anchor=west, align=left, text width=4cm] at (5,-3.2) {  
            \textbf{Pre-correction \ac{rmswfe}}:\\ 2.8729 rad\\[0.2em]
            \textbf{After correction \ac{rmswfe} }\\ \ac{zrnet}: 1.1087 rad
        };
        \node[inner sep=0pt] (fig3) at (0,-7) {  
            \includegraphics[width=0.8\textwidth]{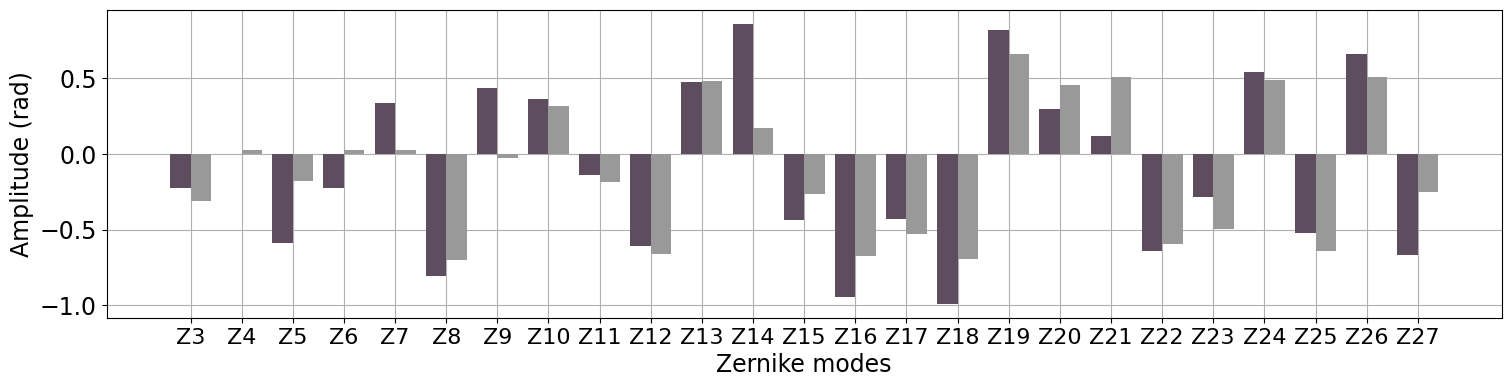}
        };
        
        \node[anchor=north west, font=\bfseries] at ($(fig3.north west) + (-0.2,0.1)$) {(c)};

        \node[anchor=west, align=left, text width=4cm] at (5,-6.8) {  
            \textbf{Pre-correction \ac{rmswfe}}: \\2.7990 rad\\[0.2em]
            \textbf{After correction \ac{rmswfe}}\\ \ac{zrnet}: 1.3023 rad
        };
  
        \node[inner sep=0pt] (fig4) at (0,-10.5) {  
             \includegraphics[width=0.8\textwidth]{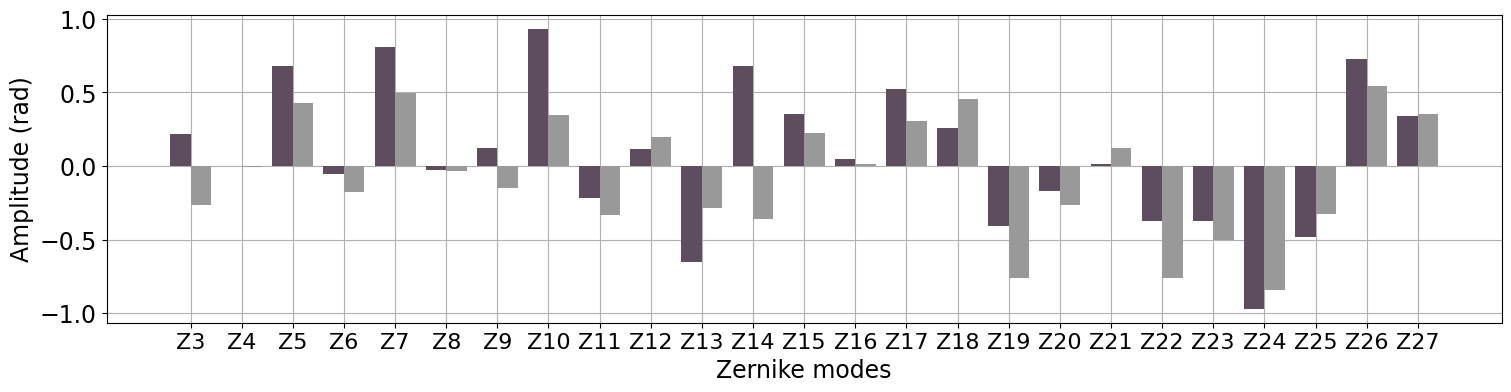}
        };

        \node[anchor=north west, font=\bfseries] at ($(fig4.north west) + (-0.2,0.1)$) {(d)};

        \node[anchor=west, align=left, text width=4cm] at (5,-10.3) {  
             \textbf{Pre-correction \ac{rmswfe}}:\\ 2.4039 rad\\[0.2em]
            \textbf{After correction \ac{rmswfe} }\\ \ac{zrnet}: 1.6020 rad
        };
        \node[inner sep=0pt] (fig5) at (0,-14) {  
            \includegraphics[width=0.8\textwidth]{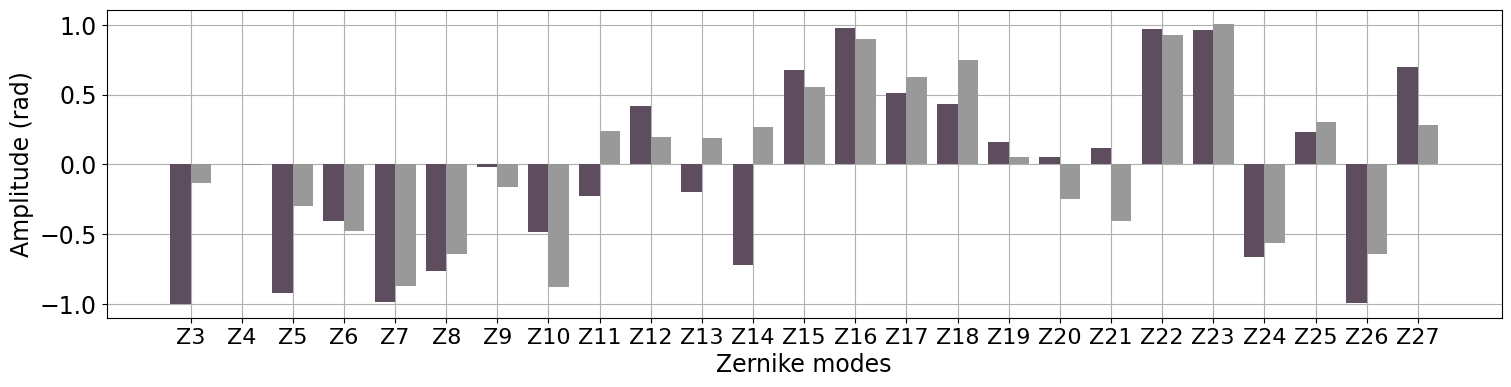}
        };
        
        \node[anchor=north west, font=\bfseries] at ($(fig5.north west) + (-0.2,0.1)$) {(e)};

        \node[anchor=west, align=left, text width=4cm] at (5,-13.8) {  
            \textbf{Pre-correction \ac{rmswfe}}: \\3.2098 rad\\[0.2em]
            \textbf{After correction \ac{rmswfe}}\\ \ac{zrnet}: 1.8864 rad
        };
        
        \node[inner sep=0pt] (fig6) at (0,-17.5) {  
            \includegraphics[width=0.8\textwidth]{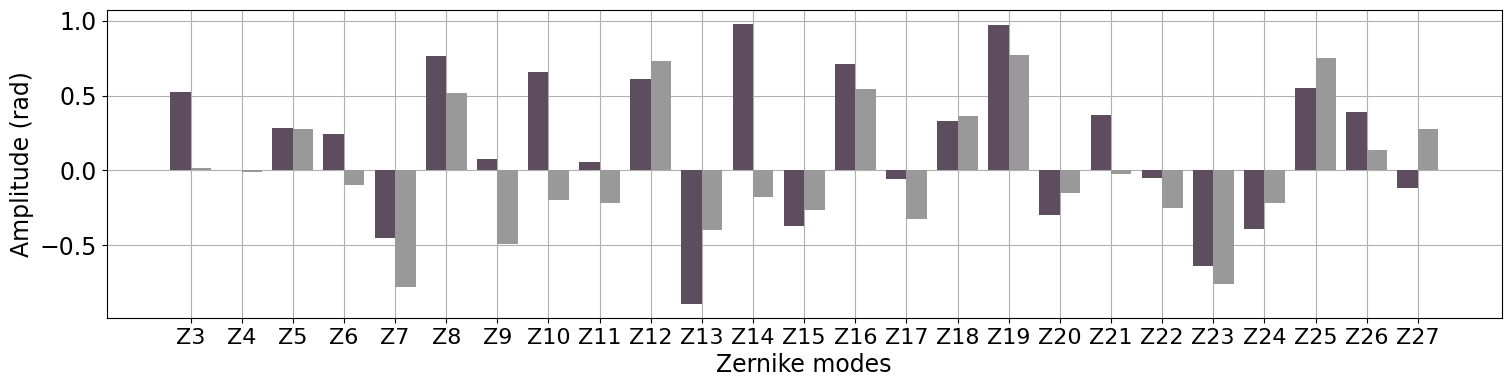}
        };
        
        \node[anchor=north west, font=\bfseries] at ($(fig6.north west) + (-0.2,0.1)$) {(f)};
        
        \node[anchor=west, align=left, text width=4cm] at (5,-17.3) {  
            \textbf{Pre-correction \ac{rmswfe}}:\\ 2.5917 rad\\[0.2em]
            \textbf{After correction \ac{rmswfe} }\\ \ac{zrnet}: 1.9825 rad
        };
    \end{tikzpicture}
    \caption{Zernike coefficients prediction on experimental \ac{psf} data comparing ground truth to ZRNet(w/o image restoration) predictions. }
    \label{fig:real_zernike}
\end{figure*}

Figure~\ref{fig:real_zernike} presents samples ordered by increasing post-correction \ac{rmswfe}. Cases with higher initial aberration severity (e, f) tend to retain larger residual errors, though the model still captures the overall coefficient profile.

\bibliographystyle{elsarticle-num}
\bibliography{bibsample}
\end{document}